\documentclass[sn-mathphys-num]{sn-jnl}% Math and Physical Sciences Numbered Reference Style 
\usepackage{graphicx}%
\usepackage{multirow}%
\usepackage{amsmath,amssymb,amsfonts}%
\usepackage{amsthm}%
\usepackage{mathrsfs}%
\usepackage[title]{appendix}%
\usepackage{xcolor}%
\usepackage{textcomp}%
\usepackage{manyfoot}%
\usepackage{booktabs}%
\usepackage{algorithm}%
\usepackage{algorithmicx}%
\usepackage{algpseudocode}%
\usepackage{listings}%
\usepackage{optidef}%
\usepackage{pifont}
\usepackage{comment}
\usepackage{hyperref}
\usepackage{float}
\usepackage{color}
\usepackage{subcaption}
\usepackage{caption}
\usepackage{mathtools}
\usepackage{multirow}
\usepackage{leftindex}

\usepackage{a4wide}

\mathtoolsset{showonlyrefs}
\usepackage{geometry}
\newtheorem{thm}{Theorem}
\newtheorem{lem}[thm]{Lemma}

\newtheorem{Definition}{Definition}

\theoremstyle{thmstyleone}%

\theoremstyle{thmstyletwo}%

\theoremstyle{thmstylethree}%

\begin{document}

\title[Article Title]{Neural Fractional Differential Equations}
%for Irregularly-sampled Data}


\author*[1]{\fnm{C.} \sur{Coelho}}\email{cmartins@cmat.uminho.pt}

\author[1]{\fnm{M. Fernanda} \sur{P. Costa}}\email{mfc@math.uminho.pt}

\author[1,2]{\fnm{L.L.} \sur{Ferrás}}\email{lferras@fe.up.pt}

\affil[1]{\orgdiv{Centre of Mathematics (CMAT)}, \orgname{University of Minho}, \orgaddress{\city{Braga}, \postcode{4710-57}, \country{Portugal}}}

\affil[2]{\orgdiv{Department of Mechanical Engineering (Section of Mathematics) and CEFT - Centro de Estudos de Fenómenos de Transporte - FEUP}, \orgname{University of Porto}, \city{Porto}, \postcode{4200-465}, \country{Portugal}}


\abstract{
  Fractional Differential Equations (FDEs) are essential tools for modelling complex systems in science and engineering. They extend the traditional concepts of differentiation and integration to non-integer orders, enabling a more precise representation of processes characterised by non-local and memory-dependent behaviours.
    This property is useful in systems where variables do not respond to changes instantaneously, but instead exhibit a strong memory of past interactions.
    Having this in mind, and drawing inspiration from Neural Ordinary Differential Equations (Neural ODEs), we propose the Neural FDE, a novel deep neural network framework that adjusts a FDE to the dynamics of data. 
    This work provides a comprehensive overview of the numerical method employed in Neural FDEs and the Neural FDE architecture. The numerical outcomes suggest that, despite being more computationally demanding, the Neural FDE may outperform the Neural ODE in modelling systems with memory or dependencies on past states, and it can effectively be applied to learn more complex dynamical systems.
}

\keywords{Neural Fractional Differential Equations, Neural Ordinary Differential Equations, Neural Networks, Time-Series, Numerical Methods.}



\maketitle

\section{Introduction}

Real systems in science and engineering, exhibit complex behaviours, often characterised by complicated dynamics and non-linear interactions. These complexities arise in various contexts, such as the interactions of molecules within a cell, the chaotic movement of turbulent flows, and the challenging task of predicting financial markets.

To predict and understand these system's behaviour efficiently, mathematical models, particularly Differential Equations (DEs), are often used, avoiding the need for costly or time-consuming experiments.

With the emergence of Neural Networks (NNs) and their impressive performance in fitting mathematical models to data, numerous studies have focused on modelling real-world systems. However, conventional NNs are designed to model \emph{functions} in a discrete way and may not be able to accurately capture the continuous dynamics observed in several systems. To overcome this limitation, Chen et al. \cite{chenNeuralOrdinaryDifferential2019a} introduced the Neural Ordinary Differential Equations (Neural ODEs), a NN architecture that adjusts an Ordinary Differential Equation (ODE) to the dynamics of a system. 

% ODEs are favoured for their simplicity and effectiveness in describing the instantaneous rates of change within a system, however, they may not adequately model systems with strong dependence on past states. To address this issue, Fractional Differential Equations (FDEs) have been introduced \cite{Podlubny}. FDEs generalise derivatives to  (non-integer) orders, allowing for modelling systems with memory and non-local interactions. By using all past states to predict current and future behaviours, they can capture fractional-order rates of change, providing a more broaden representation of a system dynamics \cite{herrmannFractionalCalculusIntroduction2014}.
ODEs are simple and effective for describing instantaneous rates of change, but may fail to model systems with strong dependence on past states, which FDEs address by using fractional derivatives to account for memory and non-local interactions \cite{herrmannFractionalCalculusIntroduction2014}.

Inspired by Neural ODEs and by the inherent memory of FDEs, in a preliminary work (a conference proceedings), we briefly proposed a novel deep learning architecture that models a FDE to the hidden dynamics of given discrete data \cite{neuralFDE}. To the best of our knowledge, this was the first time that a NN framework was proposed to fully fit a FDE to the dynamics of data, including the order of the fractional derivative.

In this work, we build upon \cite{neuralFDE} by introducing new and significant contributions. We provide: essential context on Neural ODEs and FDEs to better understand the proposed Neural FDE; explain the mechanisms and theoretical advantages of Neural FDEs, analyze the time and memory complexity of Neural FDEs, offering insights into the method's computational efficiency; enhance previous experiments by introducing additional systems; assess the experimental convergence properties and stability of Neural FDEs through various experiments, providing insights into the method's reliability.

It should be remarked that Neural ODEs and Neural FDEs are often confused with other methods in the literature that use NNs to approximate solutions of ODEs or FDEs \cite{jafarian2017artificial,pang2019fpinns}, where the ODE or FDE modelling the data dynamics is already known. In contrast, the Neural FDE (or the Neural ODE) aims to \emph{find} (under some restrictions) the FDE (or ODE) that captures the dynamics of a certain given data, representing a completely different paradigm. Note also that in this work we focus on time-series/sequential data, although Neural ODEs and Neural FDEs can used in other fields, such as, for example, image processing \cite{Cui2023}.

This paper is organised as follows. Section \ref{sec:background} presents a brief review of essential concepts such as Neural ODEs, fractional calculus and FDEs.
Section \ref{sec:NeuralFDE} presents the Neural FDE architecture along with its mathematical formulation and algorithm.
In Section \ref{sec:experiments} we evaluate the performance of the newly proposed Neural FDE. We use synthetic datasets describing two real-world systems and one real dataset to compare the performance of Neural FDE with a Neural ODE baseline.  
The paper ends with the summary of the findings and future directions in Section \ref{sec:conclusion}.
% {\color{red}Additionally, in Appendix \ref{app:solver} we provide an explanation on the impact of refining meshes and in Appendix \ref{app:experiments} we present numerical results for two additional real-world system datasets.}

\section{Background} \label{sec:background}

In this section, we provide the details needed to make this work more self-contained and bridge the gap between computer scientists and the mathematics of ODEs and FDEs.

\subsection{Neural Ordinary Differential Equations}

Inspired by Residual Neural Networks \cite{he2016deep}, in 2018 the Neural ODE architecture tailored for modelling time-series and sequential data characterised by continuous-time dynamics was proposed  \cite{chenNeuralOrdinaryDifferential2019a} (see also \cite{Massaroli}). These Neural ODEs can also be used in various applications, such as, for example, image classification, but the focus of this work is time-series and sequential data.

\begin{figure}[h]
    \centering
    \includegraphics[width=0.85\textwidth]{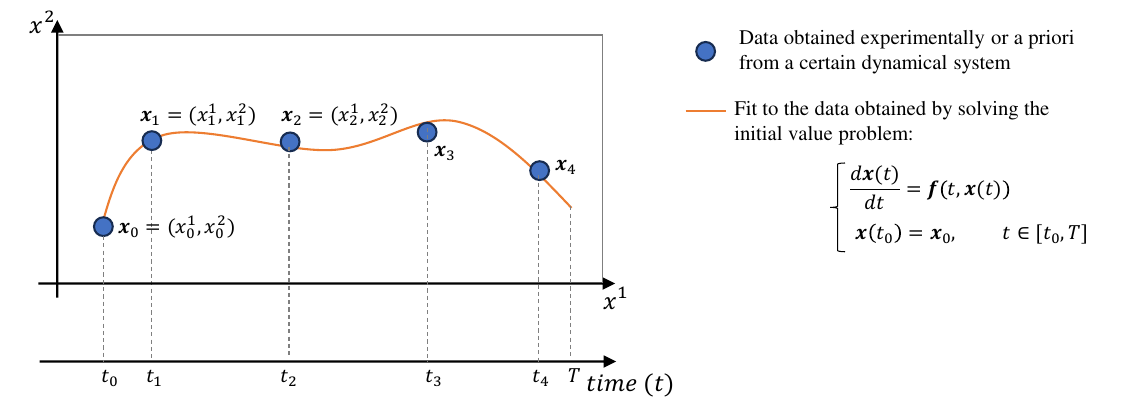}
    \caption{Fit of an ODE to data \(\{\boldsymbol{x}_0, \boldsymbol{x}_1, \boldsymbol{x}_2, \boldsymbol{x}_3, \boldsymbol{x}_4 \}\) obtained experimentally or provided by a dynamical system. The blue symbols represent the data points, while the orange line represents the fit obtained from the initial value problem shown on the right. Each vector \(\boldsymbol{x}_i\) corresponds to a specific instant \(t_i\). The initial value problem allows us to determine the behaviour of the dynamical system at any instant within the interval \([t_0, T]\).
}
    \label{fig:data}
\end{figure}

The idea behind Neural ODEs is simple, and illustrated in Fig. \eqref{fig:data}. Assume we have a collection of ordered data ($N+1$ ordered observations) \(\boldsymbol{x} = \{\boldsymbol{x}_0, \boldsymbol{x}_1, \dots, \boldsymbol{x}_{N}\}\), which represent the state of some dynamical system at discrete instants \(t_i\) over the time interval $[t_0,T]$ (with $t_N=T$). Each $\boldsymbol{x}_i= (x_i^1,x_i^2,\dots, x_i^d)  \in \mathbb{R}^d$, \(i = 0, \dots, N\) is associated with instant \(t_i\). In Fig. \eqref{fig:data} we consider $N=4$ and $d=2$ for illustrative purposes.

We assume that the given data can be modelled by the initial value problem in Eq. \eqref{eq:physicalProb}. This allows us to determine the behaviour of the dynamical system at any instant within the interval \([t_0, T]\).

\begin{equation} \label{eq:physicalProb}
    \begin{cases}
    \dfrac{d \boldsymbol{x}(t)}{dt} = \boldsymbol{f}(t, \boldsymbol{x}(t)) \\
     \boldsymbol{x}(t_0) = \boldsymbol{x}_0,\quad t \in[t_0,T].
    \end{cases}
\end{equation}
The problem is that neither the solution \(\boldsymbol{x}(t)\in\mathbb{R}^d\) nor the function \(\boldsymbol{f}(t, \boldsymbol{x}(t)): \mathbb{R}\times\mathbb{R}^d\rightarrow\mathbb{R}^d\) are known (the ODE may be linear, nonlinear, etc). Furthermore, it may seem presumptuous to assume that the given data can be modelled by this specific initial value problem. However, the rationale for selecting this particular type of differential equation will be provided in subsection \eqref{ResNet}.

\smallskip

Neural ODEs provide a viable solution to approximate the initial value problem \eqref{eq:physicalProb} using only the data \(\boldsymbol{x} = \{\boldsymbol{x}_0, \boldsymbol{x}_1, \dots, \boldsymbol{x}_{N}\}\) (the ground truth in our Neural ODE). 

\begin{figure}[h]
    \centering
    \includegraphics[width=1.0\textwidth]{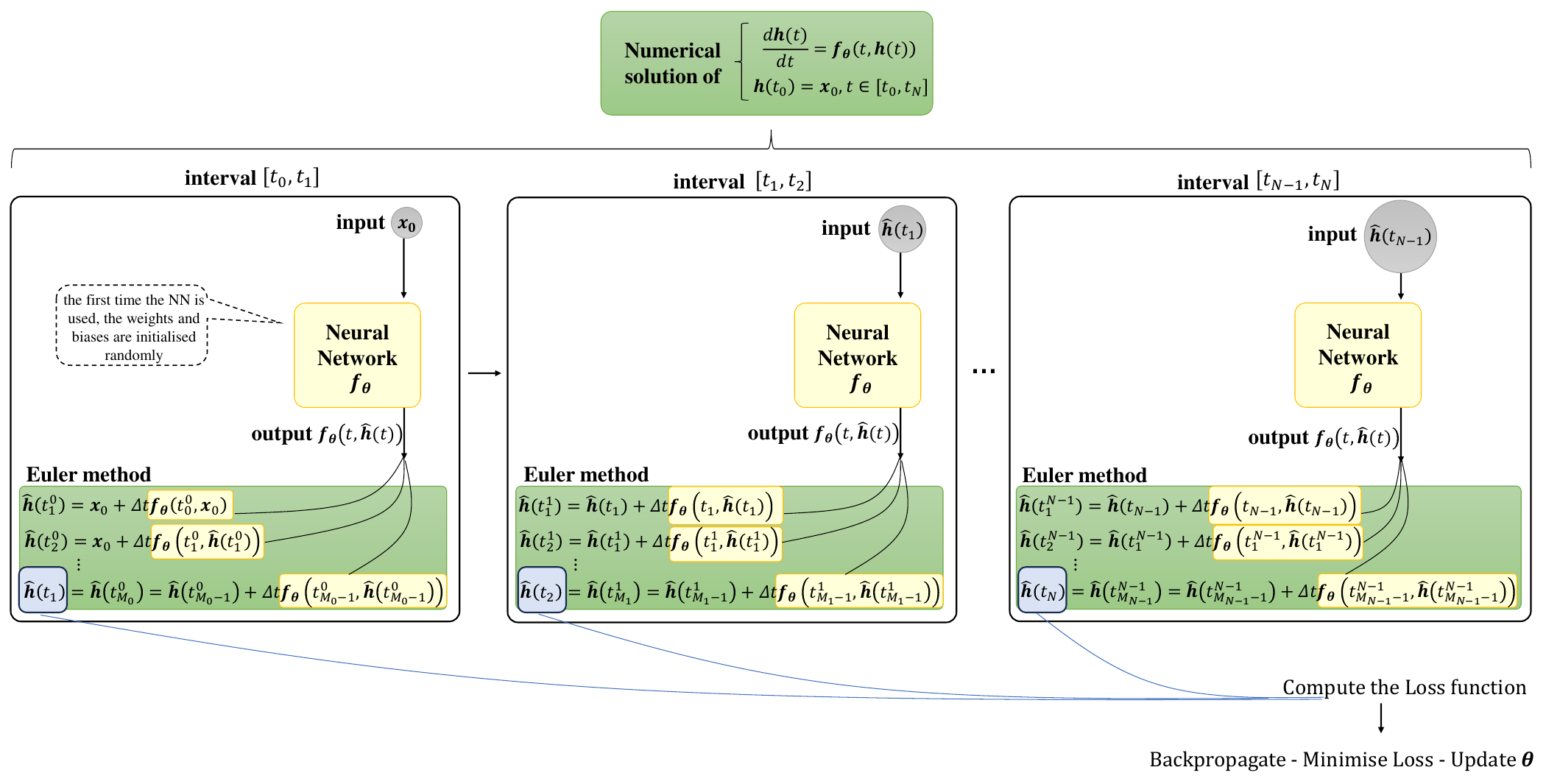}
    \caption{Schematic of a Neural ODE iteration. Note that along the sequence of figures (left to right) the NN \(\boldsymbol{f}_{\boldsymbol{\theta}}(t, \boldsymbol{h}(t))\) doesn't change. }
    \label{fig:neuralODE}
\end{figure}

Let \(\boldsymbol{h}(t)\) be an approximation of \(\boldsymbol{x}(t)\). In the Neural ODE framework (Eq. \eqref{eq:NeuralODE}) the left-hand side remains a derivative operator \(\frac{d \boldsymbol{h}(t)}{dt}\), but the right-hand side's analytical expression $(\boldsymbol{f}(t, \boldsymbol{x}(t)))$ is replaced by a NN (denoted by \(\boldsymbol{f}_{\boldsymbol{\theta}}(t,\boldsymbol{h}(t))\)), where \(\boldsymbol{\theta}\) represents the weights and biases of the network that will learn the function \(\boldsymbol{f}(t,\boldsymbol{h}(t))\) based only on the observations \(\boldsymbol{x}\) (Fig. \eqref{fig:neuralODE}),

\begin{equation} \label{eq:NeuralODE}
    \begin{cases}
     \dfrac{d \boldsymbol{h}(t)}{dt} = \boldsymbol{f}_{\boldsymbol{\theta}}(t,\boldsymbol{h}(t)), \\
     \boldsymbol{h}(t_0) = \boldsymbol{x}_0  \quad t \in[t_0,T].
    \end{cases}
\end{equation}

We can say that the Neural ODE consists of two main components: 
a numerical ODE solver that provides the numerical solution to Eq. \eqref{eq:NeuralODE}, and 
the neural network \(\boldsymbol{f}_{\boldsymbol{\theta}}(t,\boldsymbol{h}(t))\), which is supplied to the numerical solver at each evaluation within the solver. Since the output of the Neural ODE model at time \( t \) is not an exact value but a numerical approximation, we denote this solution as \( \boldsymbol{\hat{h}}(t) \) (Fig. \eqref{fig:neuralODE}).

\begin{figure}[h]
    \centering
    \includegraphics[width=0.9\textwidth]{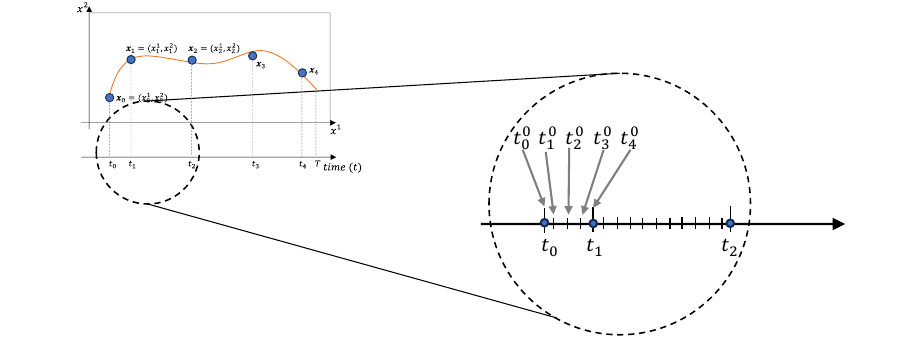}
    \caption{Example of a typical mesh used in the numerical solution of Eq. \eqref{eq:NeuralODE} for each interval $[t_i,t_{i+1}]$, $i=0,\dots, N-1$, where $t_i$ is the time of observation $\boldsymbol{x}_i$.}
    \label{fig:meshODE}
\end{figure}

To illustrate the Neural ODE, we assume that the numerical method used to solve the initial value problem \eqref{eq:NeuralODE} is the explicit Euler method \cite{Griffiths2010}. Here we follow the idea adopted in \cite{chenNeuralOrdinaryDifferential2019a} where a mesh is defined for each interval $[t_i,t_{i+1}]$, $i=0,\dots, N-1$. Therefore, given the initial condition $\boldsymbol{h}(t_0) = \boldsymbol{x}_0$, a (uniform) mesh $\{t_m^i = m\Delta t_i: m = 0, 1, . . . , M_i\}$ on an interval $[t_i,t_{i+1}]$ with some integer $M_i$ and $\Delta t := (t_{i+1}-t_i) /M_i$, we compute the numerical solution as (for the interval $[t_0,t_1]$),

\begin{align*} 
\boldsymbol{\hat{h}}(t_1^0)&=\boldsymbol{x}_0+\Delta t \boldsymbol{f}_{\boldsymbol{\theta}}(t_0^0,\boldsymbol{x}_0)\\
\boldsymbol{\hat{h}}(t_2^0)&=\boldsymbol{\hat{h}}(t_1^0)+\Delta t \boldsymbol{f}_{\boldsymbol{\theta}}(t_1^0,\boldsymbol{\hat{h}}(t_1^0))\\
&\vdots\\
\boldsymbol{\hat{h}}(t_{M_0}^0)&=\boldsymbol{\hat{h}}(t_{M-1})+\Delta t \boldsymbol{f}_{\boldsymbol{\theta}}(t_{M-1},\boldsymbol{\hat{h}}(t_{M-1}).
\end{align*}
Note that $t_0^0=t_0$ and $t_{M_0}^0=t_1$, as illustrated in Fig. \eqref{fig:meshODE}. For the interval \([t_1, t_2]\), we have the mesh points \(t_0^1, t_1^1, t_2^1, t_3^1, \dots, t_{M_1}^1\), and this process is repeated for all intervals of observations until we reach the last interval \([t_{N-1}, t_N]\). Note that \(M_i\) may vary from one interval to another, especially when we have irregular data as in Fig. \eqref{fig:data}.

Fig. \eqref{fig:neuralODE} illustrates an iteration of the Neural ODE model. For each interval $[t_i, t_{i+1}]$, we employ the Euler method to compute the numerical solution at the mesh points within that interval. This computation is only possible because a NN ($\boldsymbol{f}_{\boldsymbol{\theta}}$) is fed to the solver giving information about the unknown function $\boldsymbol{f}(t, \boldsymbol{x}(t))$.

To compute the optimal parameters \(\boldsymbol{\theta}\) of the NN, \(\boldsymbol{f}_{\boldsymbol{\theta}}(t, \boldsymbol{h}(t))\), one must backpropagate and minimise the loss function \(\mathcal{L}(\boldsymbol{\theta})\) (which in this case is based on the Mean Squared Error (MSE)) over the dataset of observations \(\{\boldsymbol{x}_i\}_{i=1}^{N}\):

\begin{equation}
\mathcal{L}(\boldsymbol{\theta}) = \frac{1}{N} \sum_{i=1}^{N} || \boldsymbol{\hat{h}}(t_i) - \boldsymbol{x}_i ||_2^2.
\end{equation}

In the backpropagation (done through automatic differentiation) we obtain the variation of the loss with respect to the weights and biases, represented as $\frac{\partial \mathcal{L}}{\partial \boldsymbol{\theta}}$. This derivative, $\frac{\partial \mathcal{L}}{\partial \boldsymbol{\theta}}$, is then used in the minimisation process to update $\boldsymbol{\theta}$ with optimal values and begin a new Neural ODE iteration. After achieving the desired accuracy in minimising the loss function, the model can be used to make predictions for any \(t \in [t_0, T]\) or even go beyond $T$ (extrapolation). To make these predictions, we only need to use the ODE solver once. 

\smallskip

Note that in \cite{chenNeuralOrdinaryDifferential2019a}, a different but well-known technique is used to \emph{backpropagate}. The authors also use both adaptive and fixed-step ODE solvers. When the step size is explicitly specified $\Delta t$, the discretization takes place for each sub-interval of observations $[t_i, t_{i+1}]$ with the specified step size yielding $(t_{i+1} - t_i)/\Delta t$ time steps (the case illustrated in Fig. \eqref{fig:meshODE}). 
On the other hand, when using an adaptive-step step solver, the discretization also occurs for each sub-interval $[t_i, t_{i+1}]$, but the step size is not predetermined, instead it is dynamically adjusted based on the local solution behaviour (local gradient) \cite{torchdiffeq}. The refinement of the mesh inside each sub-interval allows for the solver to compute the solution with higher accuracy at an increased computational cost.

% For a detailed explanation on the implications of the use of different step sizes see Appendix \ref{app:solver}.

Different numerical solvers can be used to obtain the numerical solution of \eqref{eq:NeuralODE} \cite{Griffiths2010}, for the time being, we will refer to a numerical solver as ODESolve. Assuming $i = 1, \dots, N$ and that that $t_{N}$ corresponds to $T$, each state $\boldsymbol{h}(t_i)$ is then numerically given by,

\begin{equation}
    \boldsymbol{\hat{h}}(t_i) = \text{ODESolve}(\boldsymbol{f}_{\boldsymbol{\theta}}, \boldsymbol{x}_0, \{t_1, \dots, t_N\}).
\end{equation}
Later in this work, when presenting the Neural FDE model (see Subsection \eqref{subsec:numericalsolver}), it will become clear that the mesh points used in the ODE solver do not need to coincide with the observation times associated with \(\boldsymbol{x}_i\), as illustrated in Fig. \eqref{fig:meshODE}.

It is important to note that, as shown in \cite{Dupont,Massaroli}, the approach used for Neural ODEs in \cite{chenNeuralOrdinaryDifferential2019a} has some limitations and hybrid approaches can be used \cite{Wohlleben2022}. Additionally, the concept of fitting a differential equation to the dynamics of data using the adjoint method is not entirely new.

\medskip

\noindent \textbf{Remark:} Neural ODEs can be used in various applications, such as, for example, image classification. In this scenario, an image is fed into a neural network (e.g., a convolutional neural network), which transforms the data before passing it to the NN $\boldsymbol{f}_{\boldsymbol{\theta}}$ (see Fig. \eqref{fig:neuralODE}). When solving numerically the ODE, we are only interested in the solution $\boldsymbol{\hat{h}}(t_N)$ (solutions at intermediate instants are not needed). This solution is then processed through a softmax layer to predict whether the image contains, for example, a dog or a cat.

\subsubsection{Neural ODEs and Residual Neural Networks} \label{ResNet}

Neural ODEs can be viewed as a continuous version of the (discrete) Residual Networks (ResNet) \cite{he2016deep,Haber2018,Haber2017,E2017,pmlr-v80-lu18d,Ruthotto2019,Dupont,Massaroli}. In Residual Networks, we consider the following transformation of a hidden state from layer \(t\) to layer \(t + 1\):

\begin{equation}
\label{eq:resnet}
    \boldsymbol{h}_{t+1} = \boldsymbol{h}_{t} + \boldsymbol{f}_t(\boldsymbol{h}_{t}, \boldsymbol{\theta}_{t}),
\end{equation}
where \(\boldsymbol{h}_{t} \in \mathbb{R}^d\) is the hidden state at layer \(t\), \(\boldsymbol{\theta}_{t}\) represents the parameters of the network determined by the learning process (the weights and biases), and \(\boldsymbol{f}_t : \mathbb{R}^d \rightarrow \mathbb{R}^d\) is a differentiable function. 

Assuming we have a finite number of layers, for the residual forward problem presented in \eqref{eq:resnet} to be stable, it is recommended to control the variation of \(\boldsymbol{f}_t(\boldsymbol{h}_{t}, \boldsymbol{\theta}_{t})\) across iterations. Therefore, in \cite{Haber2017, Haber2018} (see also \cite{pmlr-v80-lu18d}), the authors proposed an improved version of \eqref{eq:resnet} by introducing a positive constant \(\delta\), enabling control over the variation of $\boldsymbol{f}_t$. The smaller the \(\delta\), the more control we have over the variation of $\boldsymbol{f}_t$:

\begin{equation}
\label{eq:resnet2}
    \boldsymbol{h}_{t+1} = \boldsymbol{h}_{t} + \delta \boldsymbol{f}_t(\boldsymbol{h}_{t}, \boldsymbol{\theta}_{t}).
\end{equation}

This equation can be rewritten as:

\begin{equation}
\label{eq:resnetDiscrete}
    \dfrac{\boldsymbol{h}_{t+1} - \boldsymbol{h}_{t}}{\delta} = \boldsymbol{f}_t(\boldsymbol{h}_{t}, \boldsymbol{\theta}_{t}),
\end{equation}
and taking the limit \(\delta \rightarrow 0\), we obtain the following ODE,

\begin{equation}
\label{eq:resnetCont}
    \dfrac{d \boldsymbol{h}(t)}{dt} = \boldsymbol{f}(t,\boldsymbol{h}(t), \boldsymbol{\theta}(t)).
\end{equation}
defined over a certain time interval $t\in[t_0,T]$ with $T>t_0$. Note that while it is indeed true that the parameter $\delta$ provides a means to regulate the stability of the forward problem \eqref{eq:resnet2}, the stability is not solely determined by $\delta$ but also influenced by the variations in the weights – refer to \cite{Haber2017,Haber2018} for a comprehensive explanation.

 Eq. \eqref{eq:resnetCont} extends the discrete nature of the Residual Network, Eq. \eqref{eq:resnet2}, to a continuous model with the capability to derive the solution or state for any given moment within the specified time interval. We now have an \emph{infinite} number of layers represented by the continuity of $\boldsymbol{h}(t)$, and we also have weights and biases ($\boldsymbol{\theta}$(t)) that may or may not depend on time. For a rigorous discussion on this generalisation from discrete to continuous, see also \cite{Dupont,Massaroli}. For other architectures involving differential equations and NN please consult \cite{Ruthotto2019}.

\subsection{Fractional Calculus}

Fractional calculus deals with differential and integral operators of non-integer orders, despite the name \emph{fractional}, and its origin can be traced back to the 17th century \cite{Diethelm2010,Podlubny, Ross1975, machado2011recent, herrmannFractionalCalculusIntroduction2014}. Several definitions of fractional derivatives have been proposed in the literature by different authors, however, in this subsection we will restrict ourselves to the Riemann-Liouville and Caputo definitions, which are often used in applications to physics and engineering.

It all starst with the Fundamental Theorem of Calculus stating that:

\begin{thm}
 Let $f:[a,b]\rightarrow\mathbb{R}$ be a continuous function, and let $F:[a,b]\rightarrow\mathbb{R}$ be defined by
\begin{equation}
F(t)=\intop_{a}^{t}f(s)ds.\label{LPspace-1}
\end{equation}
\noindent Then, $F$ is differentiable and $F'=\frac{dF}{dt}=f.\label{LPspace-1-1}$
\end{thm}

Let \( D \) represent the differential operator that maps a function \( f \) to its derivative, denoted as \( Df(t) = f'(t) \). Similarly, let \( J_a \) be the integral operator that maps a function \( f(t) \) to its antiderivative, provided the integration is feasible over the compact interval \([a,\,b]\):

\begin{equation}
J_{a}f(t)=\int_{a}^{t}f(s)\,ds,\,\,x\in[a,\,b].
\label{LPspace-1-2}
\end{equation}

\noindent These operators can be extended to \( n \)-fold iterations:

\begin{equation}
D^{n}f(t) = \frac{d}{dt} \cdots \frac{d}{dt} \frac{df}{dt} = D^{1}D^{n-1}f(t),\label{LPspace-1-1-1}
\end{equation}

\begin{equation}
J_{a}^{n}f(t) = \int_{a}^{t} \cdots \int_{a}^{t} \int_{a}^{t} f(s)\,ds = J_{a}^{1}J_{a}^{n-1}f(t).\label{LPspace-1-2-1}
\end{equation}

The following lemma provides a method to express the \emph{n-fold} integral using a single integral symbol:

\begin{lem}
\label{lema_integral} Let \( f \) be Riemann integrable on \([a,\,b]\). Then, for \( a \leq x \leq b \)
and \( n \in \mathbb{N} \), we have,
\begin{equation}
J_{a}^{n}f(t) = \frac{1}{(n-1)!} \intop_{a}^{t} (t-s)^{n-1} f(s)\,ds.
\label{LPspace-1-2-1-1}
\end{equation}  
\end{lem}

To extend the previous integral to non-integer orders, one simply replaces \((n-1)!\) with an operator that generalises the factorial to non-integer values. This can be done using the Gamma function,
\begin{equation}
\Gamma(z) = \int_{0}^{\infty} t^{z-1} e^{-t} \, \mathrm{d}t, \label{12eq:cont-1-2}
\end{equation}
noting that \(\Gamma(n) = (n-1)!\) for \( n \in \mathbb{N} \). The Riemann-Liouville fractional integral is then defined as follows:

\begin{Definition} Let \( n \in \mathbb{R}_{+} \) and \( J_{a}^{n} \) be the operator defined
on \( L_{1}[a,\,b] \) by,
\begin{equation}
J_{a}^{n} f(t) = \frac{1}{\Gamma(n)} \intop_{a}^{t} (t-s)^{n-1} f(s)\,ds, \,\, x \in [a,\,b],
\label{LPspace-1-2-1-1-3}
\end{equation}
\noindent then \( J_{a}^{n} \) is called the Riemann-Liouville fractional integral operator of order \( n \).
\end{Definition}

The fractional derivative is obtained by differentiating the fractional integral defined above. In the classical case (integer orders), we have the following lemma:

\begin{lem}\label{lema_exchange}
 Let \( m,\, n \in \mathbb{N} \) with \( m > n \), and let \( f \) be a function with a continuous \( n^{\text{th}} \) derivative on the interval \([a,\,b]\). Then,
\begin{equation}
D^{n} f = D^{m} J_{a}^{n-n} f. \label{LPspace-1-2-1-1-1}
\end{equation}
\end{lem}

This lemma can be generalised to define the Riemann-Liouville fractional derivative \cite{Ross1975}:

\begin{Definition} Let \( \alpha \in \mathbb{R}_{+} \) and \( m = \lceil \alpha \rceil \). The Riemann-Liouville
fractional derivative of order \( \alpha \) (\( _{a}^{R}D_{t}^{\alpha} f \))
is given by,
\begin{equation}
_{a}^{R}D_{t}^{\alpha} f(t) = D^{m} J_{a}^{m-\alpha} f(t) = \frac{D^{m}}{\Gamma(m-\alpha)} \intop_{a}^{t} (t-s)^{m-\alpha-1} f(s)\,ds,\label{eq:integral_noninteger-1}
\end{equation}
\noindent where for the case \( n = 0 \) we have \( _{a}^{R}D_{t}^{0} := I \).
\end{Definition}

This definition of the fractional derivative may exhibit properties that are less desirable. For example, the Riemann-Liouville derivative of a constant is not zero. By rearranging the order of integration and differentiation, this \emph{less desirable characteristic} can be eliminated, leading to a new definition of the fractional derivative proposed by M. Caputo \cite{caputo1967}:

\begin{Definition}\label{Caputo} Let \( \alpha \in \mathbb{R}_{+} \), \( m = \lceil \alpha \rceil \), and \( D^{m} f(t) \in L_{1}([a,\,b]) \). The Caputo fractional derivative of order \( \alpha \) (\( _{a}^{C}D_{t}^{\alpha} f \))
is given by 
\begin{equation}
_{a}^{C}D_{t}^{\alpha} f(t) = J_{a}^{m-\alpha} D^{m} f(t) = \frac{1}{\Gamma(m-\alpha)} \intop_{a}^{t} (t-s)^{m-\alpha-1} D^{m} f(s)\,ds. \label{eq:integral_noninteger-1-2-1}
\end{equation}
\end{Definition}
This definition is more suited for modelling physical processes, since the initial conditions are based on integer-order derivatives. Therefore, in this work, we focus only on Caputo fractional derivatives of order \(0 < \alpha < 1\), and, for simplicity, we set \(a = 0\), resulting in the following expression:
\begin{equation}
    \leftindex^C_0 {D}^\alpha_t f(t) = \dfrac{1}{\Gamma(1 - \alpha)} \int_0^t (t-s)^{ - \alpha} f'(s) \, ds.
\end{equation}

To illustrate the non-locality of this fractional derivative, we consider the following initial value problem,

\begin{equation} \label{eq:ivpODE}
    \begin{cases}
      \leftindex^C_0 {D}^\alpha_t z(t) = f(t,z(t)), & t \in[0,T] \\
      z(0) = z_0
    \end{cases}
\end{equation}
where \( T \) is an arbitrary positive constant. Under certain hypotheses on the solution \( z(t) \) and the function \( f(t,z(t)) \), this initial value problem is equivalent to \cite{Diethelm2010}:

\begin{equation}
\label{eq:output1dFDE}
 z(t) = z(0) + \dfrac{1}{\Gamma(\alpha)} \int_0^t (t-s)^{ \alpha-1} f(s,z(s)) \, ds.
\end{equation}

When \(\alpha = 1\), we recover the classical case:

\begin{equation}
\label{eq:output_classical}
 z(t) = z(0) + \int_0^t f(s,z(s)) \, ds.
\end{equation}

In this classical case, consider two instants \( t_1 \) and \( t_2 \) with \( t_2 > t_1 \) and \( t_1, t_2 \in [0,T] \). We can easily compute \( z(t_1) \) if we know the initial condition and the function \( f(s,z(s)) \):

\begin{equation}
\label{eq:output_t1}
 z(t_1) = z(0) + \int_0^{t_1} f(s,z(s)) \, ds.
\end{equation}

For \( t = t_2 \), we have:

\begin{equation}
\label{eq:output_t2}
 z(t_2) = z(0) + \int_0^{t_2} f(s,z(s)) \, ds = z(0) + \int_0^{t_1} f(s,z(s)) \, ds + \int_{t_1}^{t_2} f(s,z(s)) \, ds = z(t_1) + \int_{t_1}^{t_2} f(s,z(s)) \, ds.
\end{equation}

This means that we can compute the solution at \( t_2 \) knowing only the solution at \( t_1 \) and the variation of \( f(s,z(s)) \) in the interval \( [t_1, t_2] \). Thus, the solution at \( t_2 \) does not require information from the interval \( [0, t_1) \), making $D^1 z(t)$ a local derivative operator.

When \( 0 < \alpha < 1 \), following similar ideas as in the classical case, we have:

\begin{equation}
\label{eq:nonlocal}
 z(t_2) = z(t_1) + \underline{\dfrac{1}{\Gamma(\alpha)} \int_0^{t_1} \left( (t_2 - s)^{\alpha-1} - (t_1 - s)^{\alpha-1} \right) f(s,z(s)) \, ds} + \dfrac{1}{\Gamma(\alpha)} \int_{t_1}^{t_2} (t_2 - s)^{\alpha-1} f(s,z(s)) \, ds.
\end{equation}

This indicates that the operator is non-local since, to compute the solution at any instant \( t \), we always use the full information from the past (see the underlined term in Eq. \eqref{eq:nonlocal}) \cite{Diethelm2010, Barros2021}.

\section{Neural Fractional Differential Equations}
\label{sec:NeuralFDE}
\begin{figure}[h]
    \centering
    \includegraphics[width=1\textwidth]{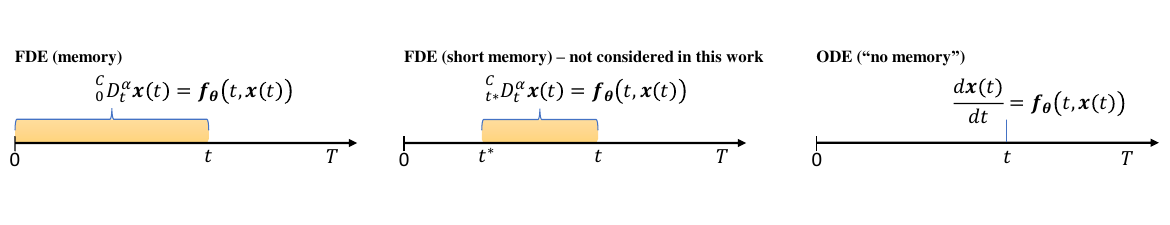}
    \caption{Memory in FDEs (left) and ODEs (right). Schematic of the computation of the derivative (fractional or classical) at instant $t$. The case of short memory (center) is presented for illustrative purposes, aiming to facilitate a clear understanding of this phenomenon.}
    \label{fig:memory}
\end{figure}

Having in mind the concepts of Neural ODEs \cite{chenNeuralOrdinaryDifferential2019a} and the role of fractional calculus in Neural systems \cite{Lundstrom2008}, we extend Neural ODEs to Neural FDEs, where the classical derivative is replaced by a Caputo fractional derivative of order $\alpha$ with $0<\alpha<1$ (when $\alpha=1$ we recover the Neural ODE). This means that our discrete set of ordered data \(\boldsymbol{x} = \{\boldsymbol{x}_0, \boldsymbol{x}_1, \dots, \boldsymbol{x}_{N}\}\) ($N+1$ observations), may be continuously approximated by a dynamical system of the form,

\begin{equation} \label{eq:FDEproblem}
    \begin{cases}
     \leftindex^C_0 {D}^{\alpha}_t  \boldsymbol{x}(t) = \boldsymbol{f}(t,\boldsymbol{x}(t)),\\
     \boldsymbol{x}(t_0) = \boldsymbol{x}_0,  \quad t \in[t_0,T], 
    \end{cases}
\end{equation}
where, once again, $\boldsymbol{x}(t)$ and $\boldsymbol{f}(t,\boldsymbol{x}(t))$ are not known. As in Neural ODEs, we will approximate this initial value problem by a Neural FDE:

\begin{equation} \label{eq:NeuralFDE}
    \begin{cases}
     \leftindex^C_0 {D}^{\alpha_{\boldsymbol{out}}}_t  \boldsymbol{h}(t) = \boldsymbol{f}_{\boldsymbol{\theta}}(t,\boldsymbol{h}(t)),  \\
     \boldsymbol{x}(t_0) = \boldsymbol{x}_0, \quad t \in[t_0,T].
    \end{cases}
\end{equation}
The idea is to take advantage of the inherent ability of fractional derivatives to capture long-term memory (schematically illustrated in Fig. \ref{fig:memory}), and improve the modelling of time-series and sequential data characterised by continuous-time dynamics \cite{neuralFDE} (note that Neural FDEs are not restricted to time-series and may be also used in different applications).

Comparing Eq. \eqref{eq:NeuralODE} (Neural ODE) with Eq. \eqref{eq:NeuralFDE} (Neural FDE), we realise that, besides the different derivative operator, there is an extra parameter, \(\alpha\), which is the order of the derivative. When modelling physical processes with FDEs, \(\alpha\) is always a parameter that raises some concerns. This is because it is often difficult to transition from molecular physics to the macro scale and obtain a \emph{correct physical value} for \(\alpha\). Therefore, in Neural FDEs, the parameter \(\alpha\) is learned by another neural network, \(\alpha_{\boldsymbol{\phi}}\), with parameters \(\boldsymbol{\phi}\) (this is why in Eq. \eqref{eq:NeuralFDE} we see the output of the NN $\alpha_{\boldsymbol{out}}$ instead of $\alpha$). This approach makes the process of adjusting Eq. \eqref{eq:NeuralFDE} to the training data completely independent of any user intervention, giving the Neural FDE the freedom to find the best possible fit.

\begin{figure}[h]
    \centering
    \includegraphics[width=0.95\textwidth]{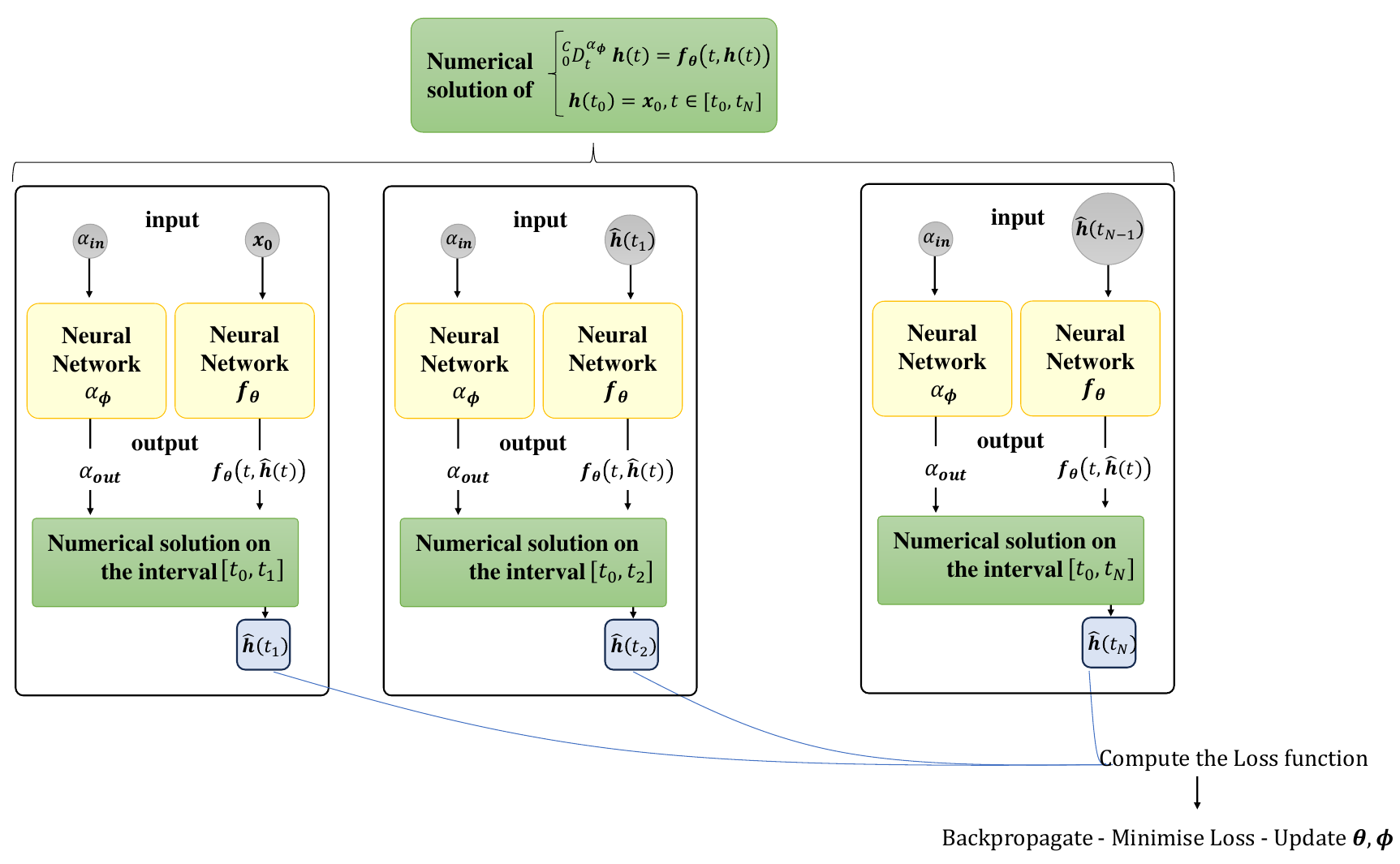}
    \caption{Schematic of a Neural ODE iteration. Note that along the sequence of figures (left to right) the NNs \(\boldsymbol{f}_{\boldsymbol{\theta}}(t, \boldsymbol{h}(t))\) and $\alpha_{\boldsymbol{\phi}}$ don't change. Note also that in the next iteration, the output $\alpha_{\boldsymbol{out}}$ will be used as the input for $\alpha_{\boldsymbol{\phi}}$ throughout the entire new iteration.}
    \label{fig:neuralFDE}
\end{figure}

Fig. \eqref{fig:neuralFDE} shows a schematic of an iteration of the Neural FDE model. In this model, two distinct neural networks pass information to the FDE solver. The NN \(\boldsymbol{f}_{\boldsymbol{\theta}}\) learns the behaviour of the analytical expression \(\boldsymbol{f}(t, \boldsymbol{h}(t))\) using as inputs \(\boldsymbol{x}_0\) and $\boldsymbol{\hat{h}}(t_i)$ (a different input for each time interval considered in the numerical solver).
The neural network \(\alpha_{\boldsymbol{\phi}}\) learns \(\alpha\) using an \(\alpha_{\boldsymbol{in}}\) value provided by the user as input and outputs a single \(\alpha\) value to be used in the numerical solver (\(\alpha_{\boldsymbol{\phi}}\)).

% For both NNs, we consider only one hidden layer to facilitate easy understanding of the results and for comparison purposes. {\color{red}The number of nodes is .... for \(\boldsymbol{f}_{\boldsymbol{\theta}}\) and ... for \(\alpha_{\boldsymbol{\phi}}\).}

To compute the optimal parameters \(\boldsymbol{\theta}\) and $\boldsymbol{\phi}$  of the two NNs, one must minimise the loss function \(\mathcal{L}(\boldsymbol{\theta},\phi)\), over the dataset \(\{\boldsymbol{x}_i\}_{i=1}^{N-1}\), that is,

\begin{mini}|l|[0]
    {\boldsymbol{\theta} \in \mathbb{R}^{n_\theta}, \boldsymbol{\phi}\in\mathbb{R}^{n_{\phi}}}{\mathcal{L}(\boldsymbol{\theta},\boldsymbol{\phi}) = \frac{1}{N} \sum_{i=1}^{N} || \boldsymbol{\hat{h}}(t_i) - \boldsymbol{x}_i ||_2^2}
    {\label{eq:constrained}}
    {}
    \addConstraint{\boldsymbol{\hat{h}}(t_i)= \text{FDESolve}}{ (\alpha_{\boldsymbol{\phi}},\boldsymbol{f}_{\boldsymbol{\theta}}, \boldsymbol{x}_0, \{t_0, t_1, \dots, t_N\})},i=1,\dots,N
    % \addConstraint{\boldsymbol{\hat{x}}(t_0)}{=\boldsymbol{x}_0},
\end{mini}
where $\text{FDESolve}(\boldsymbol{f}_{\boldsymbol{\theta}}, \boldsymbol{x}_0, \{t_0, t_1, \dots, t_N\})$ represents the numerical solver used to obtain the numerical solution $\boldsymbol{\hat{h}}(t)$ for each instant $t_i$. 

Note that, as with Neural ODEs, we can obtain the numerical solution for any instant \( t \in [t_0, t_N] \), and not just at the discrete points \( \{t_0, t_1, \dots, t_N\} \) (see subsection \ref{subsec:numericalsolver} for more details). However, it is important to highlight a significant difference in the numerical solver for FDEs when compared to the one used for Neural ODEs. To compute the solution at any instant \( t_i \), one must always use the information from the entire interval \( [t_0, t_i] \), which increases the computational burden (see subsection \ref{subsec:numericalsolver} and Fig. \ref{fig:neuralFDE}).

 As in the case of Neural ODEs, in each iteration, we proceed from left to right (Fig. \eqref{fig:neuralFDE}) to compute the approximations $\{\boldsymbol{\hat{h}}(t_i)\}_{i=1}^{N}$. Then, we backpropagate to obtain the variation of the loss with respect to the weights and biases, for both NNs, leading to $\frac{\partial \mathcal{L}}{\partial \boldsymbol{\theta}}$ and $\frac{\partial \mathcal{L}}{\partial \boldsymbol{\phi}}$. These derivatives are then used to update $\boldsymbol{\theta}$ and $\boldsymbol{\phi}$ with optimal values, and begin a new iteration. Details about the optimizer used in this work can be found in Section \ref{sec:experiments}, Numerical Experiments.

The algorithms for training (Algorithm \ref{alg:NeuralFDE_train}) and testing (Algorithm \ref{alg:NeuralFDE_test}) a Neural FDE are now presented:

\begin{algorithm}[H]
\caption{Neural FDE training process.}
\label{alg:NeuralFDE_train}
\begin{algorithmic}
\State \textbf{Input:} start time $t_0$, end time $t_N=T$, initial condition $\boldsymbol{x}(t_0) = \boldsymbol{x}_0$, mesh, maximum number of iterations \emph{MAXITER}; 
\State Choose $Optimiser$;
\State $\boldsymbol{f_\theta} = DynamicsNN()$;
\State $\alpha_{\boldsymbol{\phi}} = AlphaNN()$;
\State Initialise $\boldsymbol{\theta}, \boldsymbol{\phi}$;
\For{$k=1:$\emph{MAXITER}}
    \State $\alpha_{out} \leftarrow \alpha_{\boldsymbol{\phi}}$;
    \State $\{\boldsymbol{\hat{h}}(t_i)\}_{i=1\dots,N} \leftarrow \text{FDESolve}(\alpha, \boldsymbol{f_\theta}, \boldsymbol{x}_{0}, \{t_0, t_1, \dots, t_N\})$;
    \State Evaluate loss $\mathcal{L}$;
    \State $\nabla \mathcal{L} \leftarrow \text{Compute gradients of } \mathcal{L}(\boldsymbol{\theta},\boldsymbol{\phi})$;
    \State $\boldsymbol{\theta}, \boldsymbol{\phi} \leftarrow Optimiser.Step(\nabla \mathcal{L})$;
\EndFor
\State \textbf{Return:} $\boldsymbol{\theta}, \alpha_{out}$;
\end{algorithmic}
\end{algorithm}

\begin{algorithm}[H]
\caption{Neural FDE prediction process.}
\label{alg:NeuralFDE_test}
\begin{algorithmic}
\State \textbf{Input:} start time $t_0$, end time $t_f$, initial condition $\boldsymbol{x}(t_0) = \boldsymbol{x}_0$, parameters $\boldsymbol{\theta}$, order $\alpha$; 
\State Load $\boldsymbol{f_\theta}$;
 \State $\{\boldsymbol{\hat{h}}(t_i)\}_{i=1\dots,f} \leftarrow \text{FDESolve}(\alpha, \boldsymbol{f_\theta}, \boldsymbol{x}_{0}, \{t_0, t_1, \dots, t_f\})$;
\State \textbf{Return:} $\{\boldsymbol{\hat{h}}(t_i)\}_{i=1\dots,f}$;
\end{algorithmic}
\end{algorithm}

In summary, the Neural FDE is composed of three main components, as shown in Figure \ref{fig:neuralFDE}: 

\begin{itemize}
    \item A NN that adjusts the FDE dynamics, $\boldsymbol{f_\theta}$. Any arbitrary NN architecture can be used, from the simpler multi-layer perceptron to the more complex residual network (ResNet \cite{he2016deep}).
     
    \item A NN that adjusts the order of the derivative, $\alpha_{\boldsymbol{\phi}}$ \footnote{Instead of using a second NN, one may define $\alpha$ as a trainable variable.}. Similar to $\boldsymbol{f_\theta}$, the choice of architecture for this NN is flexible. However, since the objective is to adjust a single value, a straightforward multi-layer perceptron suffices. It is important to note that, since $\alpha \in (0,1)$, the value generated by $\alpha_{\boldsymbol{\phi}}$ must remain within this interval. To achieve this, a bounded activation function in the output layer is necessary. In the experiments detailed in this paper, a sigmoid activation function was used.

    \item A FDE numerical solver: several numerical methods have been introduced in the literature to solve FDEs, and in this work we implemented the Predictor-Corrector (PC) algorithm for FDEs, as outlined in  \cite{diethelmPredictorCorrectorApproachNumerical2002}. We chose this algorithm for being general, making it suitable for both linear and nonlinear problems, as well as for single and multi-term equations. Note that any FDE numerical solver can be used with Neural FDEs.
\end{itemize}

\subsection{The Predictor-Corrector numerical method for Neural FDEs} \label{subsec:numericalsolver}

The Predictor-Corrector numerical method used here for solving the Neural FDE, is based on the Adams–Bashforth–Moulton integrator \cite{Diethelm2002}. 
As shown in Eq. \eqref{eq:output1dFDE}, the Neural FDE can be written as (for ease of understanding, we consider \( t_0 = 0 \) and a scalar solution \( h(t) \)),

\begin{equation}
\label{eq:outputNFDE-PC}
  {h}(t)= {h(0)} + \frac{1}{\Gamma(\alpha)}\int_{0}^{t} (t-s)^{ \alpha_{\boldsymbol{out}}-1} {f}_{\boldsymbol{\theta}}(s,  {h}(s)) ds.
\end{equation}
Consider a uniform mesh $\{t_m = m \Delta t: m = 0, 1, . . . , M_i\}$ on an interval $[0,t_i]$ with some integer $M_i$ and $\Delta t := t_i /M_i$. Following the methodology used in Neural ODEs, a new mesh $\{t_m = m \Delta t : m = 0, 1, \ldots, M_i\}$ is defined for each interval $[0, t_i]$, where $t_i$ is the time associated with observation $x_i$. Consequently, the number of mesh elements ($M_i + 1$) increases with increasing $t_i$.

Let $\hat{h}(t_j)$ be a numerical approximation to the exact solution $h(t_j)$ (exact in the sense that the error that comes from evaluating $\boldsymbol{f_\theta}$ and $\alpha_{\boldsymbol{\phi}}$ is not taken into account), and that we already know the numerical solutions $\hat{h}(t_j)$, $j=0,1,\dots,n$. We then want to calculte the solution at time step $t_{n+1}$, by means of the equation,
\begin{equation}
\label{eq:outputNFDE}
  {h}(t_{n+1})= {h(0)} + \frac{1}{\Gamma(\alpha)}\int_{0}^{t_{n+1}} (t_{n+1}-s)^{ \alpha_{\boldsymbol{out}}-1} {f}_{\boldsymbol{\theta}}(s,  {h}(s)) ds.
\end{equation}

To approximate the integral $\int_{0}^{t_{n+1}} (t_{n+1}-s)^{\alpha_{\boldsymbol{out}}-1} f_{\boldsymbol{\theta}}(s, h(s)) \, ds$, we use a piecewise linear interpolation for $f_{\boldsymbol{\theta}}(s, h(s))$ at the mesh points $t_j$,
$j = 0, 1, 2, \ldots, n + 1$, with the interpolation denoted by $\overline{f}_{\boldsymbol{\theta}}(s, h(s))$. Consequently, we obtain a first approximation for $h(t_{n+1})$:

\begin{equation}
\label{eq:outputNFDE}
\hat{h}(t_{n+1}) \approx h(0) + \frac{1}{\Gamma(\alpha)}\int_{0}^{t_{n+1}} (t_{n+1}-s)^{\alpha_{\boldsymbol{out}}-1} \overline{f}_{\boldsymbol{\theta}}(s, h(s)) \, ds.
\end{equation}

Using some algebra, we find that \cite{Diethelm2002}:

\begin{equation}
\label{eq:corrector}
\hat{h}(t_{n+1}) = h(0) + \frac{(\Delta t)^{\alpha}}{\Gamma(\alpha+2)}\sum_{j=0}^{n+1}a_{j,n+1} f_{\boldsymbol{\theta}}(t_j, \hat{h}(t_j)) = h(0) + \frac{(\Delta t)^{\alpha}}{\Gamma(\alpha+2)} \left( \sum_{j=0}^{n} a_{j,n+1} f_{\boldsymbol{\theta}}(t_j, \hat{h}(t_j)) + f_{\boldsymbol{\theta}}(t_{n+1}, \underline{\hat{h}(t_{n+1})}) \right)
\end{equation}
where

\[
a_{j, n+1} = 
\begin{cases}
n^{\alpha+1} - (n-\alpha)(n+1)^{\alpha}, & \text{if } j=0 \\
(n-j+2)^{\alpha+1} + (n-j)^{\alpha+1} - 2(n-j+1)^{\alpha+1}, & \text{if } 1 \leq j \leq n\\
1, & \text{if } j=n+1.
\end{cases}
\]

Eq. \eqref{eq:predictor} is known as the \emph{Corrector} phase of the Predictor-Corrector algorithm, since we do not know $f_{\boldsymbol{\theta}}(t_{n+1}, \underline{\hat{h}(t_{n+1})})$ (in practice, this is only true for Eq.~\eqref{eq:FDEproblem} when the analytical expression on the right-hand side is known. In the Neural FDE, we can evaluate this function through the NN, but we should expect a poor approximation, therefore, we stick to the Predictor-Corrector scheme).

The \emph{Predictor} step is obtained by applying a quadrature rule to the integral in \eqref{eq:outputNFDE-PC}, without requiring information from the time-step $t_{n+1}$:

\begin{equation}\label{eq:predictor}
    \underline{\hat{h}(t_{n+1})} = h(0) + \sum_{j=0}^{n} b_{j,n+1} f_{\boldsymbol{\theta}}(t_j, h(t_j)),
\end{equation}
where

\[
b_{j,n+1} = \frac{(\Delta t)^{\alpha}}{\alpha} \left((n+1-j)^{\alpha} - (n-j)^{\alpha}\right).
\]

The numerical method involves computing Eqs. \eqref{eq:predictor} and \eqref{eq:corrector} in sequence. The error behaves as follows:

\[
\underset{j=1,\dots,M_i}{\max} \left| \hat{h}(t_j) - h(t_j) \right| = \mathcal{O}((\Delta t)^{1+\alpha}).
\]

In the Neural FDE framework, errors can also come from using two neural networks and the minimisation process. These errors should be included in an overall error analysis. Additionally, other numerical procedures could have been employed, but they were not tested in this study.

\begin{figure}[H]
    \centering
    \includegraphics[width=0.8\textwidth]{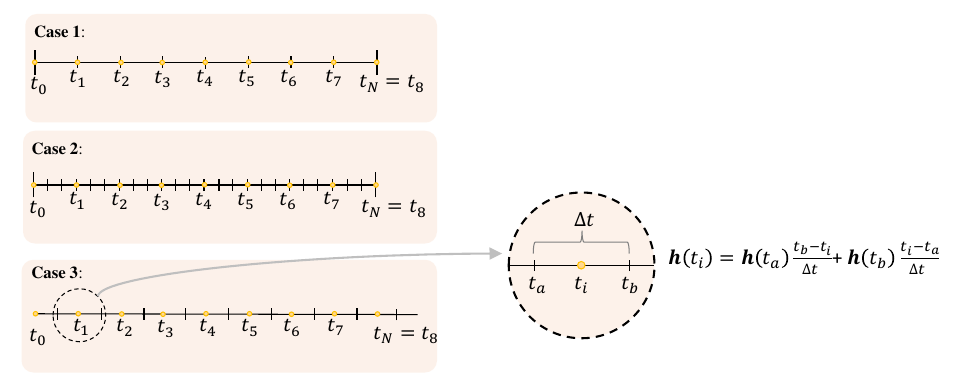}    \caption{Mesh refinement process employed in the numerical solver, along with an illustration of linear interpolation. For ease of understanding we consider a small number of observations. Case 1: the time mesh nodes (dash) used in the numerical method aligns with the input data series (symbols); Case 2: the time mesh used in the numerical method is now more refined, but for specific mesh nodes it aligns with the input data series; Case 3: the time mesh nodes used in the numerical method can be more refined or not, but they do not align with the input data series. The numerical solution is obtained for all mesh nodes considered, but for training purposes only the solutions at $t_1$, $t_2$, … are considered, being these obtained by linear interpolation (this methodology can also be used for irregularly sampled data).}
    \label{fig:mesh}
\end{figure}

As with Neural ODEs \cite{torchdiffeq}, when employing a fixed-step solver without explicitly specifying the step size \(\Delta t\) (default case), the time interval discretization occurs solely based on the time steps dictated by the time-series training data (as shown in Figure \eqref{fig:mesh} - case 1). However, when the step size is explicitly specified, the discretization takes place for the new mesh, which may (Figure \eqref{fig:mesh} - case 2) or may not (Figure \eqref{fig:mesh} - case 3) coincide with the observation instants \(t_i\) of the data \(\boldsymbol{x}\). If the mesh points do not match the training data, interpolation is used. This is illustrated in Figure \eqref{fig:mesh} - case 3.

It's important to note that, in most cases, solving FDEs of this nature results in solutions with a singularity at the origin. To tackle this issue, certain solvers incorporate a mesh refinement specifically in the vicinity of the origin \cite{Morgado0,Morgado1,Morgado2}. It's worth mentioning that, in the future, we plan to implement this graded mesh approach as a standard practice.

\subsection{Computational Cost: Neural ODE vs Neural FDE} 

% \textcolor{red}{completamente refeita - Cecília}

The most significant factor contributing to the computational cost in Neural FDEs is the memory allocation and time cost associated with the FDE solver due to the non-local nature of FDEs. Unlike ODE solvers, which rely on information from the immediate past time step, FDE solvers must retain and use values of the function at all preceding time steps to accurately compute subsequent ones \cite{diethelmPredictorCorrectorApproachNumerical2002}.

Since the primary difference between Neural ODEs and Neural FDEs lies in the solver used, we will focus our computational analysis on comparing the computational costs of the ODE and FDE solvers. Although Neural FDEs also learn the order of the derivative $\alpha$, the additional computational cost is insignificant. This is because $\alpha$ can be learnt using a simple perceptron or by adding $\alpha$ as a trainable variable, rather than employing a more complex approach involving a second NN.

In the following analysis we will consider a mesh as the one shown in Fig. \eqref{fig:mesh}-case 1.

In Neural ODEs, the computation of the next step $\boldsymbol{\hat{h}}(t_{i+1})$ at each ODE solver iteration is $\boldsymbol{\mathcal{O}}(1)$, as only the immediate previous step $\boldsymbol{\hat{h}}(t_{i})$ is needed for this computation. Let $N$ be the number of time steps used for solving the ODE, then the total computational cost is $\mathbf{\mathcal{O}}(N)$.
The storage requirement for the computed time steps is also $\boldsymbol{\mathcal{O}}(N)$.

In Neural FDEs, the computation of the next step $\boldsymbol{\hat{h}}(t_{i+1})$ at each FDE solver iteration requires the full history of past time steps $\{ \boldsymbol{\hat{h}}(t_0), \dots, \boldsymbol{\hat{h}}(t_i) \}$. As $i$ increases, the computational cost grows due to the need to perform a summation over all previous time steps, and for $t=t_N$ the total computation is $\boldsymbol{\mathcal{O}}(1) + \boldsymbol{\mathcal{O}}(2) + \dots + \boldsymbol{\mathcal{O}}\left(N(N+1)/2\right)=\boldsymbol{\mathcal{O}}(N^2).$
The storage requirement for the computed time steps remains $\boldsymbol{\mathcal{O}}(N)$.
% {\color{red} será que é mesmo??!!! Olhe que os h(t) que vai calculando, são diferentes, ou seja, quando calcula o h(t1) no intervalo [0,t2] vai ser diferentes do h(t1) calculado no intervalo [0,t3]. Isto acontece porque o kernel do integral (por exemplo ver a equaçaõ que surge logo a seguir à eq. 9) muda com o ponto final do intervalo. Será que não vamos ter na mesmo $O(N^2)$}???!!!!).

To provide a practical comparison of the computational time differences between the ODE and FDE solvers used in Neural ODE and Neural FDE, respectively, we solved an FDE and an ODE using the default \textit{Torchdiffeq} \cite{torchdiffeq} solver and the implemented PC solver:

\begin{equation}
    \leftindex^C_0 {D}^{0.6}_t y(t) + y(t) = 0, \text{ and } \dfrac{d y(t)}{dt} + y(t)=0, \,\,\, t\in(0, 20], \text{ with } y(0)=1.
\end{equation}

Each solver was run three times independently, computing 100 equally spaced time steps. The average elapsed execution time, in seconds, and the average memory used, in gigabytes (GB), is presented in Table \ref{tab:exec}.

\begin{table}[ht]
\caption{Averages of execution time and memory used for the ODE solver, default provided by \textit{Torchdiffeq}, and the implemented PC solver over three independent runs.}
\label{tab:exec}
\begin{tabular}{@{}ccc@{}}
\toprule
 & ODE Solver & FDE Solver \\ \midrule
Elapsed Time (s) & 6.23E-2 $\pm$ 1.91E-3 & 3.95E-1 $\pm$ 1.13E-2 \\
\multicolumn{1}{c}{Memory Used (GB)} & \multicolumn{1}{c}{4.22E-1 $\pm$ 5.39E-3} & \multicolumn{1}{c}{4.22E-1 $\pm$ 8.25E-5} \\ \bottomrule
\end{tabular}%
\end{table}

From Table \ref{tab:exec}, the execution time for the FDE solver is significantly higher than that of the ODE solver. Specifically, the FDE solver takes roughly $6.34$ times longer on average than the ODE solver to compute the states $\boldsymbol{\hat{h}}(t)$ for the desired time steps. Thus, the experimental results are consistent with the expected scaling behaviour. The FDE solver's execution time being an order of magnitude larger aligns qualitatively with the quadratic complexity $\boldsymbol{\mathcal{O}}(N^2)$, whereas the ODE solver's time is consistent with the linear complexity $\boldsymbol{\mathcal{O}}(N)$. Furthermore, the memory usage for both solvers is similar which corroborates with the theoretical value $\boldsymbol{\mathcal{O}}(N)$.

Therefore, in this particular case, one iteration of Neural FDE will, on average, take approximately $6$ times longer than that of Neural ODE \footnote{Note that our implementation of an FDE solver, the PC solver, has room for improvement regarding code and algorithm optimisation.}.

The characteristics of Neural ODEs make them a more efficient choice for modelling systems with simple dynamics where memory is not a critical factor. However, despite the higher computational cost being a disadvantage, it is important to consider the advantages of using Neural FDEs. These include the ability to learn more complex patterns from data and potentially achieve faster convergence.

\section{Numerical Experiments} \label{sec:experiments}

To evaluate the performance of the architectural framework proposed in this study, we conducted a comprehensive series of experiments on three different datasets. These included two synthetically generated datasets and one real-world dataset. For all experiments, the Mean Squared Error (MSE) was used to compute the loss function, denoted as $\mathcal{L}$.

\medskip

% \subsection{Synthetic Toy Datasets} \label{subsec:toy}

The synthetic toy datasets were generated by numerically solving their corresponding governing differential equations: Relaxation Oscillation Process (RO) and Population Growth (PG). For each dataset, three experiments  were conducted to comprehensively evaluate the performance of the models: 

\begin{itemize}
    \item \textit{Reconstruction:} assesses the models' capability to learn the dynamics of the training set. The evaluation is carried out using the same set for training and testing, consisting of $200$ points within the time interval $t=[0,200]$. This scenario provides insights into how well the models can reproduce the observed dynamics within the specified temporal range;
    
    \item \textit{Extrapolation:} evaluates the performance of the models in predicting unseen time horizons to simulate a future prediction scenario. The training set encompasses $200$ points in the time interval $t=[0,200]$, while the test set includes $200$ points within the extended time interval $t=[0,300]$. This experiment represents a challenging task, examining the models' generalisation capability and the ability to extrapolate beyond the observed temporal range;
    
    \item \textit{Completion:} evaluates the effectiveness of the models in missing data imputation, simulating a scenario where data are missing and need to be estimated. The training set comprises $200$ points in the time interval $t=[0,200]$, and the test set includes $400$ points within the time interval $t=[0,200]$. 
\end{itemize}

%{\color{red}Additionally, each set includes two variations: one with regularly-sampled data and another with irregularly-sampled data. The irregularly-sampled sets were generated by employing a random uniform distribution to introduce variability in the time intervals between data points. This deliberate variation in sampling patterns aims to simulate real-world scenarios where data acquisition might occur unevenly, at irregular intervals, or have missing data. Using both regularly and irregularly-sampled sets enhances the robustness of the evaluation, allowing for a more thorough examination of the models' performance across different data sampling conditions.}

Furthermore, four different fractional derivative orders, specifically $0.3, 0.5, 0.8, 0.99$, were used to generate four FDE datasets \textcolor{black}{for each system}. This selection of fractional derivative orders was made to systematically examine the capabilities and limitations of the Neural FDE in diverse scenarios. In addition, one ODE dataset was also generated to evaluate the performance of Neural FDE in modelling ODE generated data. The Neural ODE was used as a baseline in all experiments and its performance was also tested using the ODE's and FDE's datasets. 

To accommodate the randomness of the optimisation process, three independent runs were conducted for each architecture. Model assessment relied on the average Mean Squared Error ($MSE_{\text{avg}}$) and its standard deviation (std).

In all experiments we used the same architecture $\boldsymbol{f_\theta}$ for the Neural ODE and Neural FDE with: an input layer with a neuron and hyperbolic tangent (tanh) activation function; 2 hidden layers with 64 neurons and tanh activation function; and an output layer with a neuron. Since the Neural FDE has another NN, $\alpha_{\boldsymbol{\phi}}$, we use a one neuron input layer with tanh activation function, a hidden layer with 32 neurons with tanh activation function, and an output layer with a neuron and a sigmoid activation function. We always initialise the $\alpha$ value at $0.99$, in the iterative procedure. Adam optimiser \cite{kingma2014adam} was used with a learning rate of $1e-3$ and $200$ iterations on the full data sequence were performed.

All implementations were done in \textit{Pytorch} and the Neural ODE was used through \textit{Torchdiffeq} \cite{torchdiffeq}, {with the default adaptive-step solver and corresponding options}. The Predictor-Corrector FDE solver \cite{diethelmPredictorCorrectorApproachNumerical2002} was intentionally developed for this work. These solver were adapted to ensure compatibility with NNs and backpropagation through \emph{autograd}, and it is CUDA-enabled for efficient GPU use. This tailored implementation allows for the seamless integration of FDE solvers into the PyTorch framework. Additionally, the implementation of the PC solver can be used solely for solving FDEs without using NNs and with CUDA acceleration. 
To the best of our knowledge, this is the first publicly available implementation of the PC solver for FDEs in \emph{Pytorch} and being CUDA-enabled.

It is worth noting that the Neural ODE defaults to using an adaptive-step solver, whereas the solver implemented for the Neural FDE is a fixed-step solver that defaults to step-sizes equal to the sampling of the time points. 

% {\color{red} mas depois a comparação não é muito justa, certo?}

% The numerical experiments for the RC and SD systems can be found in Appendix \ref{app:RC} and Appendix \ref{app:SD}, respectively.

All computations were performed in a Google Cloud machine type g2-standard-32 with an Intel Cascade Lake 32 vCPU, 128GB RAM and a NVIDIA L4 GPU. The implementations can be found at [LINK]\footnote{available after acceptance}.

\medskip

\noindent\textbf{Remark:} Please note that the main purpose of this section on numerical experiments is not to establish superiority between the different neural network architectures. Each architecture has its own distinct advantages and disadvantages, which depend on the dynamics of the system. Overall, we will see that the Neural FDE outperforms the Neural ODE, but it's important to recognise that the mesh refinement employed during training and testing significantly impacts the final results, making a fully equitable comparison challenging.

% (see Appendix \ref{app:solver} for more details). 
Consequently, the results presented here should be viewed as a parametric examination of the behaviour of both Neural ODEs and Neural FDEs using various datasets. This serves as a means to assess and explore the methodologies outlined in the preceding sections.

\subsection{Relaxation Oscillation Process}

Relaxation oscillations are observed in various natural and engineered systems, where the dynamics alternate between periods of slow relaxation and rapid oscillations. 
Consider a system governed by the fractional-order relaxation oscillation equation:
\begin{equation}
\leftindex^C_0 {D}^\alpha_t x(t) + x(t) = 1, \quad x(0) = x_0,
\end{equation}
\noindent where $\leftindex^C_0 {D}^\alpha_t$ represents a fractional derivative operator of order $\alpha$ and $x(t)$ represents the variable that undergoes relaxation oscillations. The initial condition $x_0$ sets the initial value of $x$ at $t = 0$, determining the starting point of the oscillatory behaviour (we consider $x(t_0)=0.3$). In this fractional-order relaxation oscillation process, the variable $x(t)$ evolves over time. The system's behaviour alternates between periods of slow relaxation and rapid oscillation as it approaches the equilibrium solution $x(t) = 1$.

Considering the well known limitations in the convergence of numerical methods for fractional differential equations \cite{diethelmPredictorCorrectorApproachNumerical2002} for low $\alpha$ values, in this first case study we focused solely on datasets corresponding to fractional-order values of $\alpha = 0.8$, $0.99$, and $1$ (ODE). 
% This means that the differences between the different datasets may be smaller when compared to the case studies presented in the future subsections.

The results are organised in Table \ref{tab:RO} and Table \ref{tab:RO2} for the Neural ODE and the Neural FDE, respectively. Additionally, the learnt $\alpha$ values by the Neural FDE, for the three runs, are presented in Table \ref{tab:alphaRO}. The datasets $P_{\alpha=0.8}$, $P_{\alpha=0.99}$, and $P_{ODE}$ are constructed with their respective $\alpha$ values. For $P_{ODE}$, $\alpha$ is set to 1.

\begin{table}[h]
\caption{Performance of Neural ODE when modelling the RO regularly-sampled system ($MSE_{avg}$ $\pm$ std).}
\label{tab:RO}
\addtolength{\tabcolsep}{-0.4em}
\footnotesize
\begin{tabular}{@{}lllc@{}}
\toprule
 \bf DATASET & \multicolumn{1}{c}{\bf RECONSTRUCTION} & \multicolumn{1}{c}{\bf EXTRAPOLATION} & \bf COMPLETION \\ \midrule
$P_{\alpha=0.8}$ & 2.69E-2 $\pm$ 3.63E-2 & 6.40E-2 $\pm$ 6.42E-2 & \multicolumn{1}{l}{2.51E-2 $\pm$ 3.42E-2} \\
$P_{\alpha=0.99}$ & 9.52E-2 $\pm$ 1.05E-1 & 8.40E-2 $\pm$ 9.50E-2 & 1.13E-1 $\pm$ 1.19E-1 \\
$P_{\text{ODE}}$ & 3.47E-1 $\pm$ 3.65E-1 & 3.50E-1 $\pm$ 3.68E-1 & 3.47E-1 $\pm$ 3.65E-1 \\ \bottomrule
\end{tabular}
\end{table}

\begin{table}[h]
\caption{Performance of Neural FDE when modelling the RO regularly-sampled system ($MSE_{avg}$ $\pm$ std).}
\label{tab:RO2}
\addtolength{\tabcolsep}{-0.4em}
\footnotesize
\begin{tabular}{@{}lllc@{}}
\toprule
\bf DATASET & \multicolumn{1}{c}{\bf RECONSTRUCTION} & \multicolumn{1}{c}{\bf EXTRAPOLATION} & \bf COMPLETION \\ \midrule
$P_{\alpha=0.8}$ & 3.72E-4 $\pm$ 1.31E-4 & \multicolumn{1}{l}{1.43E-2 $\pm$ 6.32E-4} & \multicolumn{1}{l}{3.85E-4 $\pm$ 1.26E-4} \\
$P_{\alpha=0.99}$ & 3.95E-4 $\pm$ 1.44E-5 & 6.36E-4 $\pm$ 6.14E-5 & 1.12E-3 $\pm$ 3.64E-5 \\
$P_{\text{ODE}}$ & 8.80E-5 $\pm$ 1.47E-5 & 2.36E-4 $\pm$ 1.10E-4 & 1.06E-4 $\pm$ 2.90E-5 \\ \bottomrule
\end{tabular}
\end{table}

Table \ref{tab:RO} and Table \ref{tab:RO2} illustrate that the proposed Neural FDE exhibits significantly superior performance, for all datasets, compared to Neural ODE, with $MSE_{avg}$ being lower by at least two orders of magnitude. 
It is noteworthy that Neural FDE achieved a reduction of at least three orders of magnitude in error when fitting to the dataset generated by the ODE, in contrast to Neural ODE. This evidences the performance improvement over Neural ODEs. {\color{black}Although, it should be mentioned that this results were obtained with no mesh refinement in both training and testing. This justifies the fact that the Neural ODE is presenting a higher error when predicting results with a model trained with its \emph{own dataset}, $P_{ODE}$. For a more refined mesh, we would expect for the Neural ODE to perform better than the Neural FDE, when considering the dataset $P_{ODE}$. The disparity in the errors obtained will be reduced in the upcoming case studies.}

\begin{figure}[H]
    \centering
    \includegraphics[scale=0.65]{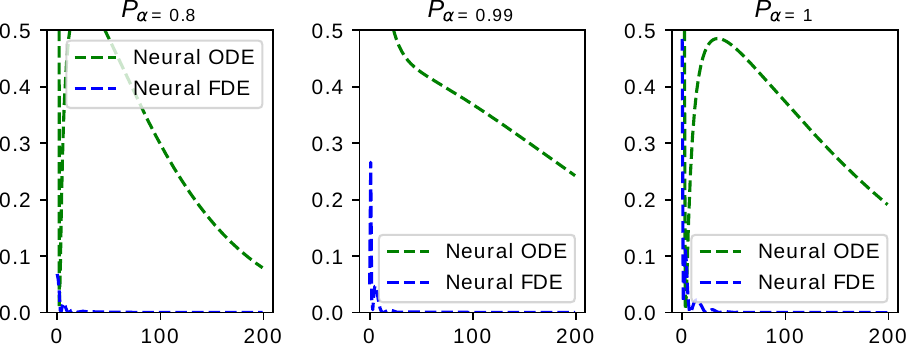}
    \caption{Evolution of the loss during training for the three datasets $P_{\alpha}=0.8,~0.99,~1$ of the RO system (loss (vertical axis) vs. iterations (horizontal axis)).}
    \label{fig:lossRO}
\end{figure}

Moreover, Figure \ref{fig:lossRO} highlights the notably faster convergence speed of Neural FDEs compared to Neural ODEs, underscoring the superiority of our proposed Neural FDE as the preferred choice. It is important to note that the 200 iterations mentioned are unrelated to the 200 data points used in constructing some datasets.

\begin{table}[h]
\caption{Adjusted $\alpha$ at each run for each dataset of the RO system.}
\label{tab:alphaRO}
\centering
\begin{tabular}{@{}lccc@{}}
\toprule
  & \multicolumn{1}{l}{$\alpha=0.8$} & \multicolumn{1}{l}{$\alpha=0.99$} & ODE    \\ \midrule
1~~~~ & 0.3885~~~~                 & 0.3451~~~~                            & 0.3015~~~~ \\
2~~~~ & 0.5477~~~~                           & 0.3923~~~~                            & 0.4431~~~~ \\
3~~~~ & 0.4472~~~~                           & 0.4684~~~~                            & 0.3692~~~~ \\ \bottomrule
\end{tabular}%
\end{table}

The $\alpha$ values learnt by the Neural FDE, Table \ref{tab:alphaRO}, show no significant patterns beyond staying relatively low, \emph{i.e.} lower than $0.5$. 

%The expected outcome was to achieve $\alpha$ values closer to those used in the training set. The observed lower values can be attributed to inadequate mesh refinement, too few training iterations, and the potential minimal impact of the $\alpha_{\phi}$ NN on the loss function (something that should be addressed in the future).

%\begin{table}[h]
%\caption{Adjusted $\alpha$ at each run for each dataset of the RO irregularly-sampled system.}
%\label{tab:alphaRO2}
%\centering
%\begin{tabular}{ccccccc}
%\hline
%  & $\alpha=0.3$ & \multicolumn{1}{l}{$\alpha=0.4$} & \multicolumn{1}{l}{$\alpha=0.5$} & \multicolumn{1}{l}{$\alpha=0.8$} & \multicolumn{1}{l}{$\alpha=0.99$} & ODE \\ \hline
%1 &              &                                  &                                  &                                  &                                   &     \\
%2 &              &                                  &                                  &                                  &                                   &     \\
%3 &              &                                  &                                  &                                  &                                   &     \\ \hline
%\end{tabular}%
%\end{table}

\subsection{Population Growth}

Consider a population of organisms that follows a fractional-order logistic growth. The population size $P(t)$ at time $t$ is governed by the following fractional-order differential equation:
\begin{equation}
 \leftindex^C_0 {D}^\alpha_t P(t) = r P(t) \left(1 - \frac{P(t)}{K}\right), \quad P(t_0) = 100, 
\end{equation}

\noindent where $\leftindex^C_0 {D}^\alpha_t$ represents the fractional derivative of order $\alpha$, $r=0.1$ is the growth rate, and $K=1000$ is the carrying capacity of the environment. Consider the initial condition $P(t_0)=100$. Please note that when employing fractional derivatives, the model parameters no longer have classical dimensions, as is the case when $\alpha=1$. These modified parameters are commonly referred to as \emph{quasiproperties}.

This equation encapsulates the intricate dynamics of the population's growth, incorporating both fractional-order derivatives and the logistic growth model. The term $r P(t) \left(1 - \frac{P(t)}{K}\right)$ characterises the population's net reproduction rate as a function of its current size relative to the environment's carrying capacity. The fractional derivative $D_t^\alpha P(t)$ introduces a nuanced temporal aspect, accounting for non-integer-order rates of change in population size over time.

\begin{figure}[H]
    \centering
    \includegraphics[scale=0.6]{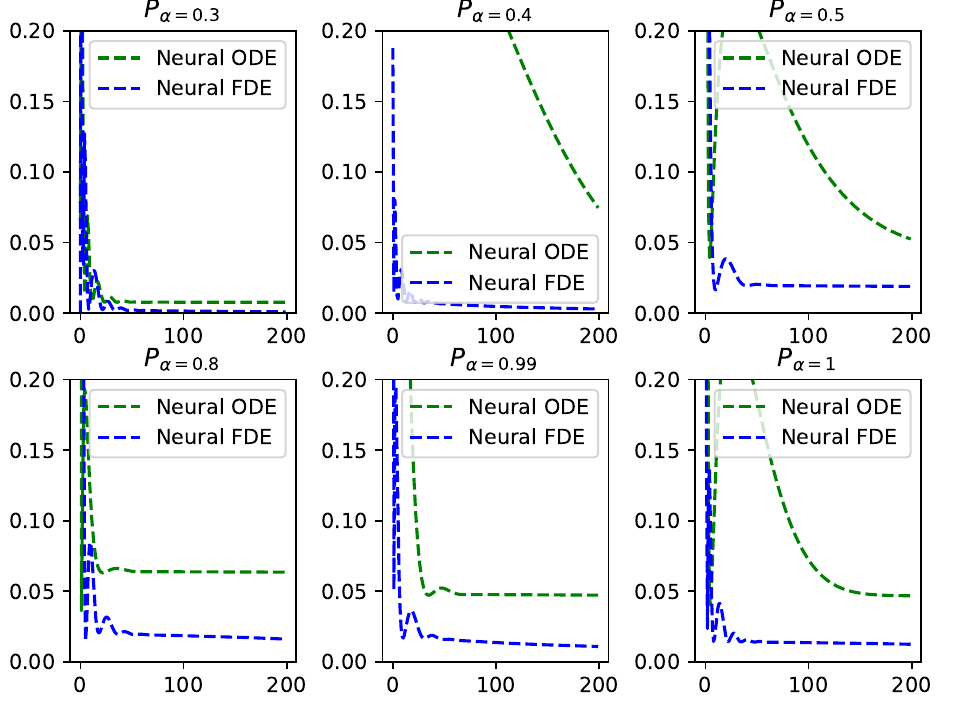}
    \caption{Evolution of the loss during training for the four datasets of the PG system using a regularly sampled dataset (loss (vertical axis) vs. iterations (horizontal axis)).}
    \label{fig:lossPG}
\end{figure}

The results are organised in Table \ref{tab:PG} and Table \ref{tab:PG2} for the Neural ODE and the Neural FDE, respectively. Additionally, the learnt $\alpha$ values by the Neural FDE, for the three runs, are shown in Table \ref{tab:alphaPG1}. To analyse and compare the convergence of Neural ODE and Neural FDE, the loss values during training were plotted in Figure \ref{fig:lossPG}.

\begin{table}[h]
\caption{Performance of Neural ODE when modelling the PG regularly-sampled system ($MSE_{avg}$ $\pm$ std).}
\label{tab:PG}
\addtolength{\tabcolsep}{-0.4em}
\footnotesize
\begin{tabular}{@{}lccc@{}}
\toprule
 \bf DATASET & \multicolumn{1}{c}{\bf RECONSTRUCTION} & \multicolumn{1}{c}{\bf EXTRAPOLATION} & \bf COMPLETION \\ \midrule
$P_{\alpha=0.3}$ & 7.70E-3 $\pm$ 2.15E-5 & 1.04E-2 $\pm$ 1.54E-5 & 7.67E-3 $\pm$ 2.15E-5 \\
$P_{\alpha=0.4}$ & 3.90E-2 $\pm$ 2.45E-2 & 4.17E-2 $\pm$ 1.91E-2 & 3.90E-2 $\pm$ 2.45E-2 \\
$P_{\alpha=0.5}$ & 4.91E-2 $\pm$ 6.26E-3 & \multicolumn{1}{c}{5.57E-2 $\pm$ 5.85E-3} & \multicolumn{1}{c}{4.92E-2 $\pm$ 6.23E-3} \\
$P_{\alpha=0.8}$ & 6.95E-2 $\pm$ 7.11E-3 & \multicolumn{1}{c}{6.74E-2 $\pm$ 1.37E-2} & \multicolumn{1}{c}{6.94E-2 $\pm$ 7.19E-3} \\
$P_{\alpha=0.99}$ & 4.58E-2 $\pm$ 1.50E-3 & 3.45E-2 $\pm$ 4.44E-4 & 4.58E-2 $\pm$ 1.50E-3 \\
$P_{\text{ODE}}$ & 8.33E-2 $\pm$ 5.19E-2 & 7.91E-2 $\pm$ 6.29E-2 & 8.34E-2 $\pm$ 5.19E-2 \\ \bottomrule
\end{tabular}
\end{table}

\begin{table}[h]
\caption{Performance of Neural FDE when modelling the PG regularly-sampled system ($MSE_{avg}$ $\pm$ std).}
\label{tab:PG2}
\addtolength{\tabcolsep}{-0.4em}
\footnotesize
\begin{tabular}{@{}lccc@{}}
\toprule
 \bf DATASET & \multicolumn{1}{c}{\bf RECONSTRUCTION} & \multicolumn{1}{c}{\bf EXTRAPOLATION} & \bf COMPLETION \\ \midrule
$P_{\alpha=0.3}$ & 1.58E-3 $\pm$ 7.34E-4 & 2.97E-3 $\pm$ 9.06E-4 & 1.96E-3 $\pm$ 7.33E-4 \\
$P_{\alpha=0.4}$ & 4.03E-3 $\pm$ 7.78E-4 & 7.36E-3 $\pm$ 1.25E-3 & 5.01E-3 $\pm$ 6.51E-4 \\
$P_{\alpha=0.5}$ & \multicolumn{1}{c}{1.99E-2 $\pm$ 1.29E-3} & \multicolumn{1}{c}{2.45E-2 $\pm$ 1.61E-3} & \multicolumn{1}{c}{2.06E-2 $\pm$ 1.30E-3} \\
$P_{\alpha=0.8}$ & \multicolumn{1}{c}{1.69E-2 $\pm$ 9.29E-3} & \multicolumn{1}{c}{1.62E-2 $\pm$ 4.83E-3} & \multicolumn{1}{c}{1.77E-2 $\pm$ 9.55E-3} \\
$P_{\alpha=0.99}$ & 1.32E-2 $\pm$ 2.05E-3 & 8.97E-3 $\pm$ 9.89E-4 & 1.45E-2 $\pm$ 1.91E-3 \\
$P_{\text{ODE}}$ & 1.15E-2 $\pm$ 1.49E-3 & 7.84E-3 $\pm$ 9.49E-4 & 1.30E-2 $\pm$ 1.37E-3 \\ \bottomrule
\end{tabular}
\end{table}

Table \ref{tab:PG} and Table \ref{tab:PG2} illustrate that the performance of both the Neural ODE and the Neural FDE is comparable in the tasks of reconstruction and completion. However, Neural FDE outperforms Neural ODE in the extrapolation task, exhibiting a lower $MSE_{avg}$, being similar for $P_{\alpha=0.5}$ and $P_{\alpha=0.8}$. This superiority underscores the significance of the memory mechanism employed by Neural FDEs, which allows models to leverage the entire historical context of the time series for making predictions. Furthermore, this memory scheme empowers Neural FDEs to capture more intricate relationships within the data, consequently achieving enhanced performance, particularly in challenging tasks involving predictions for unseen time horizons.

From Figure \ref{fig:lossPG}, it is evident that Neural FDE exhibits significantly faster convergence compared to Neural ODE, achieving lower loss values in considerably less time. This rapid convergence, coupled with its superior performance, positions Neural FDE as a highly competitive network.

\begin{table}[h]
\caption{Adjusted $\alpha$ at each run for each dataset of the PG regularly-sampled system.}
\label{tab:alphaPG1}
\centering
\begin{tabular}{@{}ccccccc@{}}
\toprule
  & \multicolumn{1}{l}{$\alpha=0.3$} & \multicolumn{1}{l}{$\alpha=0.4$} & \multicolumn{1}{l}{$\alpha=0.5$} & \multicolumn{1}{l}{$\alpha=0.8$} & \multicolumn{1}{l}{$\alpha=0.99$} & ODE    \\ \midrule
1 & 0.2792                           & 0.214                            & 0.3582                           & 0.4383                           & 0.3061                            & 0.2824 \\
2 & 0.367                            & 0.2369                           & 0.3268                           & 0.2581                           & 0.306                             & 0.2847 \\
3 & 0.2849                           & 0.2363                           & 0.3917                           & 0.411                            & 0.3631                            & 0.3374 \\ \bottomrule
\end{tabular}%
\end{table}

Once again, the $\alpha$ values learnt by Neural FDE, Table \ref{tab:alphaPG1}, show no significant patterns beyond staying relatively low, \emph{i.e.} lower than $0.4$.

At first glance, one might expect that the $\alpha$ values learnt by Neural FDEs would closely approximate the ground-truth $\alpha$ values used to generate the datasets. However, numerical experiments reveal that this is not the case. A more careful consideration of the underlying mechanisms provides insight into the reasons for this discrepancy.

Neural FDEs employ a NN $\boldsymbol{f_\theta}$ with parameters $\boldsymbol{\theta}$ to model the right-hand side of the FDE. These parameters, along with those that determine $\alpha$, are adjusted during the training process to minimise the loss function. The critical point is that, following the universal approximation theorem \cite{sonoda2017neural}, the NN $\boldsymbol{f_\theta}$ is highly adaptable and can find parameter configurations that minimise the error of the fit to the data, regardless of the specific $\alpha$ value. Consequently, there is not a unique $\alpha$ that will minimise the error, leading to multiple valid $\alpha$ values that achieve similar performance in fitting the data.

\subsubsection{Mesh refinement} \label{subsubsec:mesh}
% Selecting the appropriate $\Delta t$ poses a significant challenge due to the trade-off between accuracy of the solution and computational cost. Moreover, the optimal $\Delta t$ may vary throughout the solution domain, complicating the task of selecting a single value that adequately balances accuracy and efficiency across the entire computation. 
% To remove the need for selecting an appropriate fixed step size, adaptive-step solvers were introduced (these are only available in this work for the Neural ODE). Similar to Algorithm \ref{alg:fixedStepWithStepSize}, adaptive-step solvers divide $t$ into sub-intervals, however, $\Delta t$ is computed and dynamically adjusted during the integration process, based on the local behaviour of the solution. 

\begin{figure}[H]
    \centering
    \includegraphics[scale=0.38]{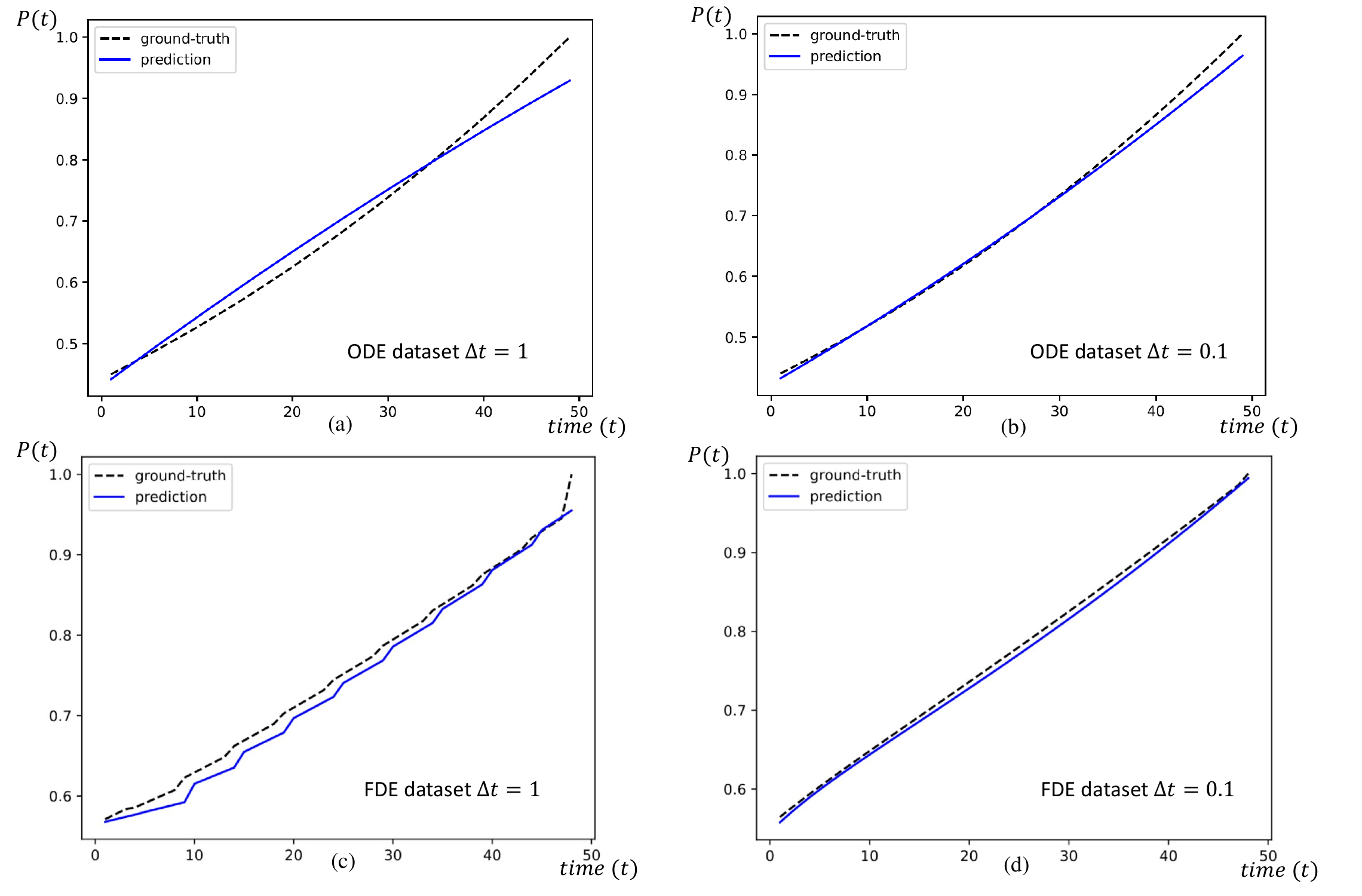}
    \caption{Mesh refinement for the PG case. (a) and (b) show the fit obtained for a ODE dataset with $\Delta t=1$ and $\Delta t=0.1$, respectively. (c) and (d) show the fit obtained for a FDE dataset with $\Delta t=1$ and $\Delta t=0.1$, respectively.}
    \label{fig:meshTest}
\end{figure}

To demonstrate how the choice of step size influences the results, we used the PG system to generate four datasets, each covering a smaller time domain $t=[0, 10]$ with $50$ data points. These datasets were created in two variations: two were derived by solving an ODE ($\alpha=1$) with step sizes of $1$ and $0.1$, while the other two were obtained by solving an FDE with $\alpha=0.8$ using step sizes of $1$ and $0.1$. Subsequently, we employed both a Neural ODE to learn the ODE generated datasets and a Neural FDE to fit the FDE generated datasets. We then performed an analysis of the plots depicting the ground truth and predicted curves, as illustrated in Figs \eqref{fig:meshTest}.

From Fig. \eqref{fig:meshTest}, it can be seen that smaller $\Delta t$ results in a more accurate approximation of the system's behaviour, as expected. This finer granularity also improves the fitting by both the Neural ODE and Neural FDE models.

For more refined meshes, the computational cost of the Neural FDE becomes more pronounced. Although the Neural FDE continues to provide better results, a fair comparison between the two models is only possible with a real-world dataset, as shown next.

\subsection{Real-world Dataset} \label{subsec:real}

The DJIA 30 Stock Time Series is a very popular irregularly sampled dataset available on Kaggle \cite{djia}. This dataset encapsulates the trajectory of the stock market across 13 years for multiple companies, offering four distinct categories of daily information: the opening market price of the stock, its highest and lowest recorded prices, and the volume of shares traded.

To conduct a comprehensive evaluation, we executed three distinct experiments to assess and compare the performance of the models:
\begin{itemize}
    \item \textit{Reconstruction:} the evaluation is carried out using the same set for training and testing, consisting of the first $365$ data points within the date range of $t=[2006/01/03, 2007/06/15]$ (Data1).
    \item \textit{Extrapolation:} the training set encompasses the first $365$ data points within the date range of $t=[2006/01/03, 2007/06/15]$ (Data1), while the test set includes $30$ data points within the extended time interval $t=[2007/06/18, 2007/07/30].$ 
    \item \textit{Completion:} the training set consists of $2717$ data points, comprising the first $2415$ within the date range of $t=[2006/01/03, 2015/08/07]$ and $302$ taken in the range of $t=[2015/08/10,2017/12/29]$ by omitting a data point between each triple of data (Data2). The omitted data point is used to construct the test set, evaluating the imputation of missing data. The test set comprises of $39$ data points.
\end{itemize}

A Neural ODE was employed as a baseline to demonstrate the performance of the Neural FDE. The evaluation was carried out through three independent runs, and the $MSE_{avg}$ along with the corresponding standard deviation were computed. The architectures and training details of NNs remain consistent with those used in Section \ref{subsec:real} with the exception of the usage of Euler with a step-size of $0.1$ for training Neural ODE and an equal step-size for Neural FDE.

\subsubsection{DJIA 30 Stock Time Series}

In this study, we conducted an initial experiment to contrast the performance of the proposed Neural FDE against Neural ODEs using a real-world dataset.

The experiments were conducted using data from the first company in the dataset, Altaba. The results are organised in Table \ref{tab:DJIA} for both the Neural ODE and Neural FDE models. Additionally, the learnt $\alpha$ values by the Neural FDE in the three independent runs are presented in Table \ref{tab:alphaDJIA}. The experiments were conducted using data from the first company in the dataset, Altaba. The results for both the Neural ODE and Neural FDE models are organised in Table \ref{tab:DJIA}. Additionally, the learnt $\alpha$ values by the Neural FDE in the three independent runs are presented in Table \ref{tab:alphaDJIA}. Note that although the reconstruction and extrapolation experiments were performed on networks trained using Data1, and completion was performed on networks trained using Data2, the training loss is similar, resembling the performance in the reconstruction tasks.

\begin{table}[h]
\caption{{Performance of Neural ODE and Neural FDE when modelling the DJIA irregularly-sampled dataset ($MSE_{avg}$ $\pm$ std).}}
\label{tab:DJIA}
\addtolength{\tabcolsep}{-0.4em}
%\footnotesize
\begin{tabular}{@{}cccc@{}}
\toprule
\bf MODEL & \bf RECONSTRUCTION & \bf EXTRAPOLATION & \bf COMPLETION \\ \midrule
Neural ODE & 2.97E-3 $\pm$ 3.32E-4 & 7.93E-2 $\pm$ 4.28E-3  & 3.29E-2 $\pm$ 4.27E-3 \\
Neural FDE & 2.76E-3 $\pm$ 4.56E-5 & 5.42E-2 $\pm$ 1.62E-3 & 1.97E-2 $\pm$ 3.35E-4  \\ \bottomrule
\end{tabular}
\end{table}

\begin{figure}[H]
    \centering
    \includegraphics[width=0.36\textwidth]{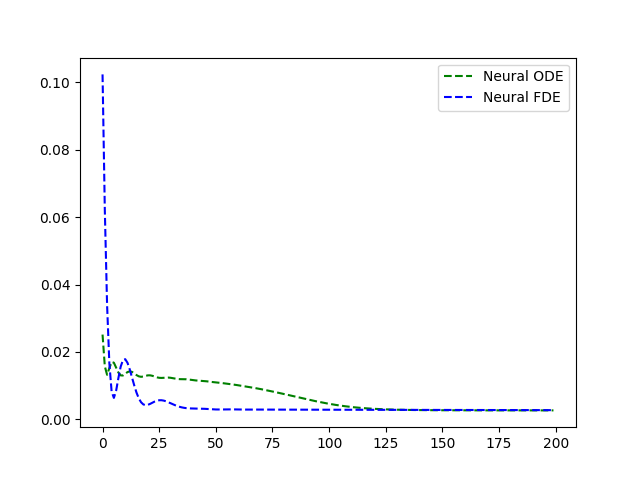}
    \includegraphics[width=0.36\textwidth]{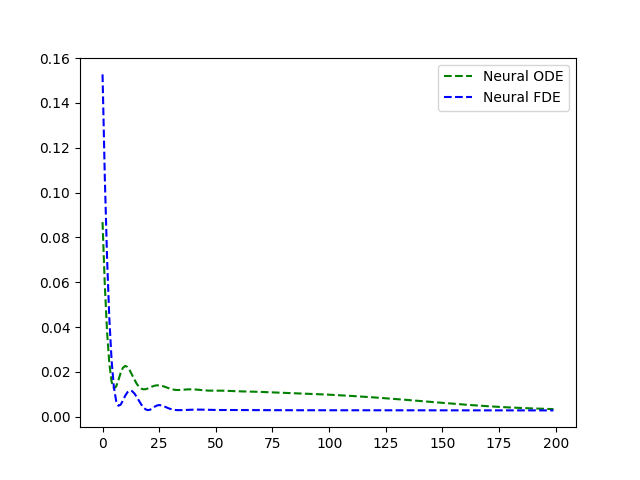}
    \caption{Evolution of the loss during training (loss (vertical axis) vs. iterations (horizontal axis)) for Data1 (on the left) and Data2 (on the right).}
    \label{fig:lossData1}
\end{figure}

Table \ref{tab:DJIA} shows that the performance of Neural FDE is comparable to the Neural ODE baseline across all three experiments. However, the loss evolution depicted in Figure \ref{fig:lossData1} indicates that Neural FDE achieves faster convergence, reaching lower loss values much earlier than Neural ODE.

Note that these are preliminary results aimed at showcasing the applicability of Neural FDE to real-world datasets as well as comparing the performance with a Neural ODE baseline. As future work, decreasing the step size, increasing the number of training iterations, and using more complex NN schemes are expected to improve performance.

\begin{table}[h]
\caption{Adjusted $\alpha$ at each run for each training dataset of the DJIA irregularly-sampled dataset.}
\label{tab:alphaDJIA}
\centering
% \begin{tabular}{@{}ccc@{}}
\begin{tabular}{ccc}
\toprule
 & \multicolumn{1}{l}{Data1} & \multicolumn{1}{l}{Data2} \\ \midrule
1~~~~~~~~ & 0.3264~~~~~~~~ & 0.3282~~~~~~~~ \\
2~~~~~~~~ & 0.3335~~~~~~~~ & 0.2933~~~~~~~~ \\
3~~~~~~~~ & 0.2985~~~~~~~~ & 0.3458~~~~~~~~ \\ \bottomrule
\end{tabular}%
\end{table}

The $\alpha$ values learnt during training for the two datasets, Table \ref{tab:alphaDJIA}, are consistent between runs. However, this consistency does not carry significant meaning since $\boldsymbol{f_\theta}$ is a universal approximator capable of adapting to any $\alpha$.

\newpage

\section{Discussion and Conclusions} \label{sec:conclusion}

Having in mind the concepts of Neural ODEs \cite{chenNeuralOrdinaryDifferential2019a} and the role of fractional calculus in Neural systems \cite{Lundstrom2008}, we extend Neural ODEs to Neural FDEs. These feature a Caputo fractional derivative of order $\alpha$ (with $0<\alpha<1$) on the left-hand side, representing the variation of the state of a dynamical system, and a neural network ($\boldsymbol{f_\theta}$) on the right-hand side. When $\alpha=1$, we recover the Neural ODE. The parameter $\alpha$ is learned by another neural network, $\alpha_{\boldsymbol{\phi}}$, with parameters $\boldsymbol{\phi}$. This approach allows the Neural FDE to adjust to the training data independently, enabling it to find the best possible fit without user intervention.

During the Neural FDE training, we employ a specialised numerical solver for FDEs to compute the numerical solution. Subsequently, a loss function is used to compare the numerically predicted outcomes with the ground truth. Using \emph{autograd} for backpropagation, we adjust the weights and biases within the neural networks $\boldsymbol{f_\theta}$ and $\alpha_{\boldsymbol{\phi}}$ to minimise the loss.

% In the study by Chen et al. \cite{chenNeuralOrdinaryDifferential2019a}, the explanation regarding the numerical methods for solving differential equations was somewhat unclear. Therefore, in this work, we provide a clear explanation of how mesh refinement can be performed and how this refinement can be applied to handle irregularly sampled data. This explanation is extended to Neural FDEs, where the Predictor-Corrector numerical method is employed.

To assess the performance of the architectural framework proposed in this study, a comprehensive series of experiments was carried out on a variety of datasets. First, two synthetic toy datasets were generated by numerically solving their corresponding differential equations. These datasets were specifically designed to model well-established systems governed by differential equations: Relaxation Oscillation Process (RO) and Population Growth (PG).

The experiments demonstrate that Neural FDE exhibits either significantly better or similar performance, as measured by $MSE_{avg}$, compared to Neural ODE. However, even in cases where the produced models demonstrate similar performance, Neural FDE showcases substantially faster convergence, achieving lower loss values in significantly fewer iterations than Neural ODE. Although, it should be mentioned that this results were obtained with no mesh refinement in both training and testing (only one mesh refinement case was considered). This justifies the fact that in some cases the Neural ODE presents a higher error when predicting results with a model trained with its \emph{own generated dataset}. For a more refined mesh, we would expect for the Neural ODE to perform better than the Neural FDE in this particular case. 
Note also that in some cases the computational times of the Neural FDE are prohibitive, and therefore, in the future, these computations should be optimised.

\bigskip

Although the Neural FDE presented in this work behaves well for the different case studies considered, there is plenty of room for improvements:

\begin{itemize}
    \item The use of FDEs needs the computation of the entire history of the dynamical system at every time step, requiring the storage of all variables. Additionally, we employed \emph{autograd} for backpropagation. This results in a significantly higher computational cost compared to Neural ODEs.

    \item When computing the loss function, we consider the weights and biases for the NN $\boldsymbol{f_\theta}$, as well as weights and biases for the NN $\alpha_{\boldsymbol{\phi}}$, optimising the order of the fractional derivative. Our numerical results indicate that this loss function does not yield optimal orders for the fractional derivative. Therefore, in future work, we should explore a different loss function that prioritises minimising the error while also optimising the order of the fractional derivative.

    \item While the study by Chen et al. \cite{chenNeuralOrdinaryDifferential2019a} does not explicitly address it, Neural ODEs exhibit stability issues. These problems are likewise present in Neural FDEs and stem from significant variations in weight matrices as they evolve in time. In the future, it is advisable to enhance the model's stability by incorporating constraints within the loss function. These constraints would enable better control over the smoothness of the weight matrix evolution.

       \item The Predictor-Corrector numerical method, employed to solve the FDE \cite{diethelmPredictorCorrectorApproachNumerical2002}, exhibits a convergence order of $\mathcal{O}((\Delta t)^p)$, where $\Delta t$ denotes the time-step and $p=min(2,1+\alpha)$. In simpler terms, this implies that lower values of $\alpha$ can yield convergence orders of 1. It's worth noting that the dependence of the convergence order on $\alpha$ is a common characteristic. To address this, one potential solution is the use of more robust numerical methods or the implementation of graded meshes, especially in the vicinity of the singularity at $t=0$.
\end{itemize}

\bmhead{Acknowledgements}

The authors acknowledge the funding by Fundação para a Ciência e Tecnologia (Portuguese Foundation for Science and Technology) through CMAT projects UIDB/00013/2020 and UIDP/00013/2020 and the funding by FCT and Google Cloud partnership through projects CPCA-IAC/AV/589164/2023 and CPCA-IAC/AF/589140/2023.

\noindent C. Coelho would like to thank FCT for the funding through the scholarship with reference 2021.05201.BD.

This work is also financially supported by national funds through the FCT/MCTES (PIDDAC), under the project 2022.06672.PTDC - iMAD - Improving the Modelling of Anomalous Diffusion and Viscoelasticity: solutions to industrial problems.

\bibliography{000-library}


\begin{thebibliography}{32}
\ifx \bisbn   \undefined \def \bisbn  #1{ISBN #1}\fi
\ifx \binits  \undefined \def \binits#1{#1}\fi
\ifx \bauthor  \undefined \def \bauthor#1{#1}\fi
\ifx \batitle  \undefined \def \batitle#1{#1}\fi
\ifx \bjtitle  \undefined \def \bjtitle#1{#1}\fi
\ifx \bvolume  \undefined \def \bvolume#1{\textbf{#1}}\fi
\ifx \byear  \undefined \def \byear#1{#1}\fi
\ifx \bissue  \undefined \def \bissue#1{#1}\fi
\ifx \bfpage  \undefined \def \bfpage#1{#1}\fi
\ifx \blpage  \undefined \def \blpage #1{#1}\fi
\ifx \burl  \undefined \def \burl#1{\textsf{#1}}\fi
\ifx \doiurl  \undefined \def \doiurl#1{\url{https://doi.org/#1}}\fi
\ifx \betal  \undefined \def \betal{\textit{et al.}}\fi
\ifx \binstitute  \undefined \def \binstitute#1{#1}\fi
\ifx \binstitutionaled  \undefined \def \binstitutionaled#1{#1}\fi
\ifx \bctitle  \undefined \def \bctitle#1{#1}\fi
\ifx \beditor  \undefined \def \beditor#1{#1}\fi
\ifx \bpublisher  \undefined \def \bpublisher#1{#1}\fi
\ifx \bbtitle  \undefined \def \bbtitle#1{#1}\fi
\ifx \bedition  \undefined \def \bedition#1{#1}\fi
\ifx \bseriesno  \undefined \def \bseriesno#1{#1}\fi
\ifx \blocation  \undefined \def \blocation#1{#1}\fi
\ifx \bsertitle  \undefined \def \bsertitle#1{#1}\fi
\ifx \bsnm \undefined \def \bsnm#1{#1}\fi
\ifx \bsuffix \undefined \def \bsuffix#1{#1}\fi
\ifx \bparticle \undefined \def \bparticle#1{#1}\fi
\ifx \barticle \undefined \def \barticle#1{#1}\fi
\bibcommenthead
\ifx \bconfdate \undefined \def \bconfdate #1{#1}\fi
\ifx \botherref \undefined \def \botherref #1{#1}\fi
\ifx \url \undefined \def \url#1{\textsf{#1}}\fi
\ifx \bchapter \undefined \def \bchapter#1{#1}\fi
\ifx \bbook \undefined \def \bbook#1{#1}\fi
\ifx \bcomment \undefined \def \bcomment#1{#1}\fi
\ifx \oauthor \undefined \def \oauthor#1{#1}\fi
\ifx \citeauthoryear \undefined \def \citeauthoryear#1{#1}\fi
\ifx \endbibitem  \undefined \def \endbibitem {}\fi
\ifx \bconflocation  \undefined \def \bconflocation#1{#1}\fi
\ifx \arxivurl  \undefined \def \arxivurl#1{\textsf{#1}}\fi
\csname PreBibitemsHook\endcsname

%%% 1
\bibitem[\protect\citeauthoryear{Chen
  et~al.}{2018}]{chenNeuralOrdinaryDifferential2019a}
\begin{botherref}
\oauthor{\bsnm{Chen}, \binits{R.T.}},
\oauthor{\bsnm{Rubanova}, \binits{Y.}},
\oauthor{\bsnm{Bettencourt}, \binits{J.}},
\oauthor{\bsnm{Duvenaud}, \binits{D.K.}}:
Neural ordinary differential equations.
Advances in neural information processing systems
\textbf{31}
(2018)
\end{botherref}
\endbibitem

%%% 2
\bibitem[\protect\citeauthoryear{Herrmann}{2014}]{herrmannFractionalCalculusIntroduction2014}
\begin{bbook}
\bauthor{\bsnm{Herrmann}, \binits{R.}}:
\bbtitle{Fractional Calculus: An Introduction for Physicists},
\bedition{2. ed} edn.
\bpublisher{{World Scientific}},
\blocation{{New Jersey, NJ}}
(\byear{2014}).
\doiurl{10.1142/9789814551083}
\end{bbook}
\endbibitem

%%% 3
\bibitem[\protect\citeauthoryear{Coelho et~al.}{2024}]{neuralFDE}
\begin{bchapter}
\bauthor{\bsnm{Coelho}, \binits{C.}},
\bauthor{\bsnm{Costa}, \binits{M.F.P.}},
\bauthor{\bsnm{Ferrás}, \binits{L.L.}}:
\bctitle{Tracing footprints: Neural networks meet non-integer order
  differential equations for modelling systems with memory}.
In: \bbtitle{Tiny Papers @ ICLR}
(\byear{2024}).
\burl{https://openreview.net/forum?id=8518dcW4hc}
\end{bchapter}
\endbibitem

%%% 4
\bibitem[\protect\citeauthoryear{Jafarian
  et~al.}{2017}]{jafarian2017artificial}
\begin{barticle}
\bauthor{\bsnm{Jafarian}, \binits{A.}},
\bauthor{\bsnm{Mokhtarpour}, \binits{M.}},
\bauthor{\bsnm{Baleanu}, \binits{D.}}:
\batitle{Artificial neural network approach for a class of fractional ordinary
  differential equation}.
\bjtitle{Neural Computing and Applications}
\bvolume{28},
\bfpage{765}--\blpage{773}
(\byear{2017})
\end{barticle}
\endbibitem

%%% 5
\bibitem[\protect\citeauthoryear{Pang et~al.}{2019}]{pang2019fpinns}
\begin{barticle}
\bauthor{\bsnm{Pang}, \binits{G.}},
\bauthor{\bsnm{Lu}, \binits{L.}},
\bauthor{\bsnm{Karniadakis}, \binits{G.E.}}:
\batitle{fpinns: Fractional physics-informed neural networks}.
\bjtitle{SIAM Journal on Scientific Computing}
\bvolume{41}(\bissue{4}),
\bfpage{2603}--\blpage{2626}
(\byear{2019})
\end{barticle}
\endbibitem

%%% 6
\bibitem[\protect\citeauthoryear{Cui et~al.}{2023}]{Cui2023}
\begin{barticle}
\bauthor{\bsnm{Cui}, \binits{W.}},
\bauthor{\bsnm{Zhang}, \binits{H.}},
\bauthor{\bsnm{Chu}, \binits{H.}},
\bauthor{\bsnm{Hu}, \binits{P.}},
\bauthor{\bsnm{Li}, \binits{Y.}}:
\batitle{On robustness of neural odes image classifiers}.
\bjtitle{Information Sciences}
\bvolume{632},
\bfpage{576}--\blpage{593}
(\byear{2023})
\doiurl{10.1016/j.ins.2023.03.049}
\end{barticle}
\endbibitem

%%% 7
\bibitem[\protect\citeauthoryear{He et~al.}{2016}]{he2016deep}
\begin{bchapter}
\bauthor{\bsnm{He}, \binits{K.}},
\bauthor{\bsnm{Zhang}, \binits{X.}},
\bauthor{\bsnm{Ren}, \binits{S.}},
\bauthor{\bsnm{Sun}, \binits{J.}}:
\bctitle{Deep residual learning for image recognition}.
In: \bbtitle{Proceedings of the IEEE Conference on Computer Vision and Pattern
  Recognition (CVPR)}
(\byear{2016})
\end{bchapter}
\endbibitem

%%% 8
\bibitem[\protect\citeauthoryear{Massaroli et~al.}{2020}]{Massaroli}
\begin{barticle}
\bauthor{\bsnm{Massaroli}, \binits{S.}},
\bauthor{\bsnm{Poli}, \binits{M.}},
\bauthor{\bsnm{Park}, \binits{J.}},
\bauthor{\bsnm{Yamashita}, \binits{A.}},
\bauthor{\bsnm{Asama}, \binits{H.}}:
\batitle{Dissecting neural odes}.
\bjtitle{Advances in Neural Information Processing Systems}
\bvolume{33},
\bfpage{3952}--\blpage{3963}
(\byear{2020})
\end{barticle}
\endbibitem

%%% 9
\bibitem[\protect\citeauthoryear{Griffiths and Higham}{2010}]{Griffiths2010}
\begin{bbook}
\bauthor{\bsnm{Griffiths}, \binits{D.F.}},
\bauthor{\bsnm{Higham}, \binits{D.J.}}:
\bbtitle{Numerical Methods for Ordinary Differential Equations},
\bedition{2010} edn.
\bsertitle{Springer undergraduate mathematics series}.
\bpublisher{Springer},
\blocation{London, England}
(\byear{2010})
\end{bbook}
\endbibitem

%%% 10
\bibitem[\protect\citeauthoryear{Chen}{2018}]{torchdiffeq}
\begin{botherref}
\oauthor{\bsnm{Chen}, \binits{R.T.Q.}}:
torchdiffeq
(2018).
\url{https://github.com/rtqichen/torchdiffeq}
\end{botherref}
\endbibitem

%%% 11
\bibitem[\protect\citeauthoryear{Dupont et~al.}{2019}]{Dupont}
\begin{botherref}
\oauthor{\bsnm{Dupont}, \binits{E.}},
\oauthor{\bsnm{Doucet}, \binits{A.}},
\oauthor{\bsnm{Teh}, \binits{Y.W.}}:
Augmented neural odes.
Advances in neural information processing systems
\textbf{32}
(2019)
\end{botherref}
\endbibitem

%%% 12
\bibitem[\protect\citeauthoryear{Wohlleben et~al.}{2022}]{Wohlleben2022}
\begin{bbook}
\bauthor{\bsnm{Wohlleben}, \binits{M.}},
\bauthor{\bsnm{Bender}, \binits{A.}},
\bauthor{\bsnm{Peitz}, \binits{S.}},
\bauthor{\bsnm{Sextro}, \binits{W.}}:
\bbtitle{Development of a Hybrid Modeling Methodology for Oscillating Systems
  with Friction},
pp. \bfpage{101}--\blpage{115}.
\bpublisher{Springer}, \blocation{???}
(\byear{2022}).
\doiurl{10.1007/978-3-030-95470-3_8} .
\burl{http://dx.doi.org/10.1007/978-3-030-95470-3_8}
\end{bbook}
\endbibitem

%%% 13
\bibitem[\protect\citeauthoryear{Haber et~al.}{2018}]{Haber2018}
\begin{botherref}
\oauthor{\bsnm{Haber}, \binits{E.}},
\oauthor{\bsnm{Ruthotto}, \binits{L.}},
\oauthor{\bsnm{Holtham}, \binits{E.}},
\oauthor{\bsnm{Jun}, \binits{S.-H.}}:
Learning across scales---multiscale methods for convolution neural networks.
Proceedings of the AAAI Conference on Artificial Intelligence
\textbf{32}(1)
(2018)
\doiurl{10.1609/aaai.v32i1.11680}
\end{botherref}
\endbibitem

%%% 14
\bibitem[\protect\citeauthoryear{Haber and Ruthotto}{2017}]{Haber2017}
\begin{barticle}
\bauthor{\bsnm{Haber}, \binits{E.}},
\bauthor{\bsnm{Ruthotto}, \binits{L.}}:
\batitle{Stable architectures for deep neural networks}.
\bjtitle{Inverse Problems}
\bvolume{34}(\bissue{1}),
\bfpage{014004}
(\byear{2017})
\doiurl{10.1088/1361-6420/aa9a90}
\end{barticle}
\endbibitem

%%% 15
\bibitem[\protect\citeauthoryear{E}{2017}]{E2017}
\begin{barticle}
\bauthor{\bsnm{E}, \binits{W.}}:
\batitle{A proposal on machine learning via dynamical systems}.
\bjtitle{Communications in Mathematics and Statistics}
\bvolume{5}(\bissue{1}),
\bfpage{1}--\blpage{11}
(\byear{2017})
\doiurl{10.1007/s40304-017-0103-z}
\end{barticle}
\endbibitem

%%% 16
\bibitem[\protect\citeauthoryear{Lu et~al.}{2018}]{pmlr-v80-lu18d}
\begin{bchapter}
\bauthor{\bsnm{Lu}, \binits{Y.}},
\bauthor{\bsnm{Zhong}, \binits{A.}},
\bauthor{\bsnm{Li}, \binits{Q.}},
\bauthor{\bsnm{Dong}, \binits{B.}}:
\bctitle{Beyond finite layer neural networks: Bridging deep architectures and
  numerical differential equations}.
In: \beditor{\bsnm{Dy}, \binits{J.}},
\beditor{\bsnm{Krause}, \binits{A.}} (eds.)
\bbtitle{Proceedings of the 35th International Conference on Machine Learning}.
\bsertitle{Proceedings of Machine Learning Research},
vol. \bseriesno{80},
pp. \bfpage{3276}--\blpage{3285}
(\byear{2018}).
\burl{https://proceedings.mlr.press/v80/lu18d.html}
\end{bchapter}
\endbibitem

%%% 17
\bibitem[\protect\citeauthoryear{Ruthotto and Haber}{2019}]{Ruthotto2019}
\begin{barticle}
\bauthor{\bsnm{Ruthotto}, \binits{L.}},
\bauthor{\bsnm{Haber}, \binits{E.}}:
\batitle{Deep neural networks motivated by partial differential equations}.
\bjtitle{Journal of Mathematical Imaging and Vision}
\bvolume{62}(\bissue{3}),
\bfpage{352}--\blpage{364}
(\byear{2019})
\doiurl{10.1007/s10851-019-00903-1}
\end{barticle}
\endbibitem

%%% 18
\bibitem[\protect\citeauthoryear{Diethelm}{2010}]{Diethelm2010}
\begin{bbook}
\bauthor{\bsnm{Diethelm}, \binits{K.}}:
\bbtitle{The Analysis of Fractional Differential Equations: An
  Application-Oriented Exposition Using Differential Operators of Caputo Type}.
\bpublisher{Springer},
\blocation{Berlin, Heidelberg}
(\byear{2010}).
\doiurl{10.1007/978-3-642-14574-2} .
\burl{http://dx.doi.org/10.1007/978-3-642-14574-2}
\end{bbook}
\endbibitem

%%% 19
\bibitem[\protect\citeauthoryear{Podlubny}{1999}]{Podlubny}
\begin{barticle}
\bauthor{\bsnm{Podlubny}, \binits{I.}}:
\batitle{Fractional differential equations: an introduction to fractional
  derivatives, fractional differential equations, to methods of their solution
  and some of their applications}.
\bjtitle{Mathematics in science and engineering}
\bvolume{198},
\bfpage{1}--\blpage{340}
(\byear{1999})
\end{barticle}
\endbibitem

%%% 20
\bibitem[\protect\citeauthoryear{Ross}{1975}]{Ross1975}
\begin{bbook}
\bauthor{\bsnm{Ross}, \binits{B.}}:
In: \beditor{\bsnm{Ross}, \binits{B.}} (ed.)
\bbtitle{A brief history and exposition of the fundamental theory of fractional
  calculus},
pp. \bfpage{1}--\blpage{36}.
\bpublisher{Springer},
\blocation{Berlin, Heidelberg}
(\byear{1975}).
\doiurl{10.1007/BFb0067096} .
\burl{https://doi.org/10.1007/BFb0067096}
\end{bbook}
\endbibitem

%%% 21
\bibitem[\protect\citeauthoryear{Machado et~al.}{2011}]{machado2011recent}
\begin{barticle}
\bauthor{\bsnm{Machado}, \binits{J.T.}},
\bauthor{\bsnm{Kiryakova}, \binits{V.}},
\bauthor{\bsnm{Mainardi}, \binits{F.}}:
\batitle{Recent history of fractional calculus}.
\bjtitle{Communications in Nonlinear Science and Numerical Simulation}
\bvolume{16}(\bissue{3}),
\bfpage{1140}--\blpage{1153}
(\byear{2011})
\end{barticle}
\endbibitem

%%% 22
\bibitem[\protect\citeauthoryear{Caputo}{1967}]{caputo1967}
\begin{barticle}
\bauthor{\bsnm{Caputo}, \binits{M.}}:
\batitle{Linear models of dissipation whose q is almost frequency
  independent--ii}.
\bjtitle{Geophysical Journal International}
\bvolume{13}(\bissue{5}),
\bfpage{529}--\blpage{539}
(\byear{1967})
\doiurl{10.1111/j.1365-246x.1967.tb02303.x}
\end{barticle}
\endbibitem

%%% 23
\bibitem[\protect\citeauthoryear{Barros et~al.}{2021}]{Barros2021}
\begin{botherref}
\oauthor{\bsnm{Barros}, \binits{L.C.d.}},
\oauthor{\bsnm{Lopes}, \binits{M.M.}},
\oauthor{\bsnm{Pedro}, \binits{F.S.}},
\oauthor{\bsnm{Esmi}, \binits{E.}},
\oauthor{\bsnm{Santos}, \binits{J.P.C.d.}},
\oauthor{\bsnm{Sánchez}, \binits{D.E.}}:
The memory effect on fractional calculus: an application in the spread of
  covid-19.
Computational and Applied Mathematics
\textbf{40}(3)
(2021)
\doiurl{10.1007/s40314-021-01456-z}
\end{botherref}
\endbibitem

%%% 24
\bibitem[\protect\citeauthoryear{Lundstrom et~al.}{2008}]{Lundstrom2008}
\begin{barticle}
\bauthor{\bsnm{Lundstrom}, \binits{B.N.}},
\bauthor{\bsnm{Higgs}, \binits{M.H.}},
\bauthor{\bsnm{Spain}, \binits{W.J.}},
\bauthor{\bsnm{Fairhall}, \binits{A.L.}}:
\batitle{Fractional differentiation by neocortical pyramidal neurons}.
\bjtitle{Nature Neuroscience}
\bvolume{11}(\bissue{11}),
\bfpage{1335}--\blpage{1342}
(\byear{2008})
\doiurl{10.1038/nn.2212}
\end{barticle}
\endbibitem

%%% 25
\bibitem[\protect\citeauthoryear{Diethelm
  et~al.}{2002}]{diethelmPredictorCorrectorApproachNumerical2002}
\begin{barticle}
\bauthor{\bsnm{Diethelm}, \binits{K.}},
\bauthor{\bsnm{Ford}, \binits{N.J.}},
\bauthor{\bsnm{Freed}, \binits{A.D.}}:
\batitle{A {{Predictor-Corrector Approach}} for the {{Numerical Solution}} of
  {{Fractional Differential Equations}}}.
\bjtitle{Nonlinear Dynamics}
\bvolume{29}(\bissue{1}),
\bfpage{3}--\blpage{22}
(\byear{2002})
\end{barticle}
\endbibitem

%%% 26
\bibitem[\protect\citeauthoryear{Diethelm and Ford}{2002}]{Diethelm2002}
\begin{barticle}
\bauthor{\bsnm{Diethelm}, \binits{K.}},
\bauthor{\bsnm{Ford}, \binits{N.J.}}:
\batitle{Analysis of fractional differential equations}.
\bjtitle{Journal of Mathematical Analysis and Applications}
\bvolume{265}(\bissue{2}),
\bfpage{229}--\blpage{248}
(\byear{2002})
\doiurl{10.1006/jmaa.2000.7194}
\end{barticle}
\endbibitem

%%% 27
\bibitem[\protect\citeauthoryear{Ford et~al.}{2013}]{Morgado0}
\begin{barticle}
\bauthor{\bsnm{Ford}, \binits{N.J.}},
\bauthor{\bsnm{Morgado}, \binits{M.L.}},
\bauthor{\bsnm{Rebelo}, \binits{M.}}:
\batitle{Nonpolynomial collocation approximation of solutions to fractional
  differential equations}.
\bjtitle{Fractional Calculus and Applied Analysis}
\bvolume{16}(\bissue{4}),
\bfpage{874}--\blpage{891}
(\byear{2013})
\doiurl{10.2478/s13540-013-0054-3}
\end{barticle}
\endbibitem

%%% 28
\bibitem[\protect\citeauthoryear{Morgado et~al.}{2021}]{Morgado1}
\begin{barticle}
\bauthor{\bsnm{Morgado}, \binits{M.L.}},
\bauthor{\bsnm{Rebelo}, \binits{M.}},
\bauthor{\bsnm{Ferrás}, \binits{L.L.}}:
\batitle{Stable and convergent finite difference schemes on nonuniformtime
  meshes for distributed-order diffusion equations}.
\bjtitle{Mathematics}
\bvolume{9}(\bissue{16}),
\bfpage{1975}
(\byear{2021})
\doiurl{10.3390/math9161975}
\end{barticle}
\endbibitem

%%% 29
\bibitem[\protect\citeauthoryear{Ferr{\'a}s et~al.}{2019}]{Morgado2}
\begin{bchapter}
\bauthor{\bsnm{Ferr{\'a}s}, \binits{L.L.}},
\bauthor{\bsnm{Ford}, \binits{N.J.}},
\bauthor{\bsnm{Morgado}, \binits{M.L.}},
\bauthor{\bsnm{Rebelo}, \binits{M.}}:
\bctitle{A hybrid numerical scheme for fractional-order systems}.
In: \beditor{\bsnm{Machado}, \binits{J.}},
\beditor{\bsnm{Soares}, \binits{F.}},
\beditor{\bsnm{Veiga}, \binits{G.}} (eds.)
\bbtitle{Innovation, Engineering and Entrepreneurship},
pp. \bfpage{735}--\blpage{742}.
\bpublisher{Springer},
\blocation{Cham}
(\byear{2019})
\end{bchapter}
\endbibitem

%%% 30
\bibitem[\protect\citeauthoryear{Kingma and Ba}{2014}]{kingma2014adam}
\begin{botherref}
\oauthor{\bsnm{Kingma}, \binits{D.P.}},
\oauthor{\bsnm{Ba}, \binits{J.}}:
Adam: A method for stochastic optimization.
arXiv preprint arXiv:1412.6980
(2014)
\end{botherref}
\endbibitem

%%% 31
\bibitem[\protect\citeauthoryear{Sonoda and Murata}{2017}]{sonoda2017neural}
\begin{barticle}
\bauthor{\bsnm{Sonoda}, \binits{S.}},
\bauthor{\bsnm{Murata}, \binits{N.}}:
\batitle{Neural network with unbounded activation functions is universal
  approximator}.
\bjtitle{Applied and Computational Harmonic Analysis}
\bvolume{43}(\bissue{2}),
\bfpage{233}--\blpage{268}
(\byear{2017})
\end{barticle}
\endbibitem

%%% 32
\bibitem[\protect\citeauthoryear{szrlee}{2018}]{djia}
\begin{botherref}
\oauthor{\bsnm{szrlee}}:
Djia 30 stock time series
(online resource -
  \href{https://www.kaggle.com/datasets/szrlee/stock-time-series-20050101-to-20171231?select=AAPL_2006-01-01_to_2018-01-01.csv}{https://www.kaggle.com/datasets/szrlee/stock-time-series-20050101-to-20171231?select=AAPL\_2006-01-01\_to\_2018-01-01.csv},
  last accessed on 2022/09/26)
(2018)
\end{botherref}
\endbibitem

\end{thebibliography}

\end{document}